\documentclass{article} %
\usepackage{iclr2024_conference,times}

\newcommand{\pst}{\textsc{Pasta}\xspace}
\newcommand{\sr}{{\textsc{Sim2Real}}\xspace}
\newcommand{\syn}{{\textsc{Sim}}\xspace}
\newcommand{\real}{{\textsc{Real}}\xspace}
\definecolor{bluebell}{rgb}{0.64, 0.64, 0.82}
\newcommand{\std}[1]{\tiny \textcolor{black}{$\pm$#1}}

\newlength\savewidth

\definecolor{defaultcolor}{HTML}{E8E2F7}

\makeatletter
\DeclareRobustCommand\onedot{\futurelet\@let@token\@onedot}
\def\@onedot{\ifx\@let@token.\else.\null\fi\xspace}

\def\eg{\emph{e.g}\onedot} 
\def\ie{\emph{i.e}\onedot}

\def\etc{\emph{etc}\onedot} 
 
\makeatother

\newcommand{\eat}[1]{}
\newcommand{\replace}[2]{} %

\newenvironment{packed_itemize}{
	\begin{list}{\labelitemi}{\leftmargin=1em}
		\vspace{-6pt}
		\setlength{\itemsep}{0pt}
		\setlength{\parskip}{0pt}
		\setlength{\parsep}{0pt}
	}{\end{list}}

\belowdisplayskip \abovedisplayskip

\newlength{\sectionReduceTop}
\newlength{\sectionReduceBot}
\newlength{\subsectionReduceTop}
\newlength{\subsectionReduceBot}
\newlength{\abstractReduceTop}
\newlength{\abstractReduceBot}
\newlength{\captionReduceTop}
\newlength{\captionReduceBot}
\newlength{\subsubsectionReduceTop}
\newlength{\subsubsectionReduceBot}

\newlength{\eqnReduceTop}
\newlength{\eqnReduceBot}

\newlength{\horSkip}
\newlength{\verSkip}

\newlength{\figureHeight}
\usepackage{amsmath,amsfonts,bm}

\def\eqref#1{equation~\ref{#1}}

\def\1{\bm{1}}

\DeclareMathAlphabet{\mathsfit}{\encodingdefault}{\sfdefault}{m}{sl}
\SetMathAlphabet{\mathsfit}{bold}{\encodingdefault}{\sfdefault}{bx}{n}

\newcommand{\E}{\mathop{\mathbb{E}}}

\newcommand{\softmax}{\mathrm{softmax}}

\DeclareMathOperator*{\argmax}{arg\,max}

\usepackage{url}

\usepackage{xcolor}
\usepackage[pagebackref=false, breaklinks=true, letterpaper=true, colorlinks,citecolor=citecolor, linkcolor=linkcolor, bookmarks=false]{hyperref}
\definecolor{citecolor}{HTML}{0071BC}
\definecolor{linkcolor}{HTML}{ED1C24}

\usepackage[utf8]{inputenc} %
\usepackage[T1]{fontenc}    %
\usepackage{hyperref}       %
\usepackage{url}            %
\usepackage{booktabs}       %
\usepackage{amsfonts}       %
\usepackage{nicefrac}       %
\usepackage{microtype}      %
\usepackage{xcolor}         %
\newcommand{\augcal}{\textsc{AugCal}\xspace}
\newcommand{\aug}{\textsc{Aug}\xspace}
\newcommand{\calib}{\textsc{Cal}\xspace}
\definecolor{Gray}{gray}{0.9}
\definecolor{aliceblue}{rgb}{0.94, 0.97, 1.0}
\usepackage{float}
\usepackage{tikz}
\usepackage{algorithm}
\usepackage{xspace}
\usepackage[noend]{algpseudocode}
\usepackage{multirow}
\usepackage{rotating}
\usepackage{booktabs}
\usepackage{tabularx}
\usepackage{tabulary}
\usepackage{rotating}
\usepackage{adjustbox}

\usepackage{color, colortbl}
\definecolor{Gray}{gray}{0.9}

\newcommand{\md}[1]{\colorbox{red!15}{#1}}
\newcommand{\ac}[1]{\colorbox{defaultcolor}{#1}}
\usepackage{amsmath, amsthm, amssymb}
\usepackage{textcomp}
\usepackage{stmaryrd}
\usepackage{upgreek}
\usepackage{bm}
\usepackage{cases}
\usepackage{mathtools}
\usepackage{array}
\usepackage{arydshln}
\usepackage{multirow}
\usepackage{bbm}
\usepackage{etoc}  
\usepackage{wrapfig}
\usepackage{subfig}
\usepackage{pythonhighlight}
\usepackage{inconsolata}

\definecolor{city_color_0}{rgb}{0.0,0.0,0.0}
\definecolor{city_color_1}{rgb}{0.5020,0.2510,0.5020}
\definecolor{city_color_2}{rgb}{0.9569,0.1373,0.9098}
\definecolor{city_color_3}{rgb}{0.2745,0.2745,0.2745}
\definecolor{city_color_4}{rgb}{0.4000,0.4000,0.6118}
\definecolor{city_color_5}{rgb}{0.7451,0.6000,0.6000}
\definecolor{city_color_6}{rgb}{0.6000,0.6000,0.6000}
\definecolor{city_color_7}{rgb}{0.9804,0.6667,0.1176}
\definecolor{city_color_8}{rgb}{0.8627,0.8627,0.0000}
\definecolor{city_color_9}{rgb}{0.4196,0.5569,0.1373}
\definecolor{city_color_10}{rgb}{0.5961,0.9843,0.5961}
\definecolor{city_color_11}{rgb}{0.2745,0.5098,0.7059}
\definecolor{city_color_12}{rgb}{0.8627,0.0784,0.2353}
\definecolor{city_color_13}{rgb}{1.0000,0.0000,0.0000}
\definecolor{city_color_14}{rgb}{0.0000,0.0000,0.5569}
\definecolor{city_color_15}{rgb}{0.0000,0.0000,0.2745}
\definecolor{city_color_16}{rgb}{0.0000,0.2353,0.3922}
\definecolor{city_color_17}{rgb}{0.0000,0.3137,0.3922}
\definecolor{city_color_18}{rgb}{0.0000,0.0000,0.9020}
\definecolor{city_color_19}{rgb}{0.4667,0.0431,0.1255}

\newlength\myheight
\newlength\mydepth
\settototalheight\myheight{Xygp}
\theoremstyle{definition}

\title{\augcal: Improving Sim2Real Adaptation \\by Uncertainty Calibration \\on Augmented Synthetic Images}

\author{
    \vspace{1mm}\centerline{\textbf{Prithvijit Chattopadhyay$\thanks{Correspondence to PC}$ $\quad$ Bharat Goyal $\quad$ Boglarka Ecsedi $\quad$ Viraj Prabhu $\quad$ Judy Hoffman}}\\
    \centerline{Georgia Tech}\\
    \centerline{\texttt{\footnotesize \{prithvijit3,bharatgoyal,becsedi3,virajp,judy\}@gatech.edu}}
}

\iclrfinalcopy %
\begin{document}

\setlength{\abovedisplayskip}{3pt}
\setlength{\belowdisplayskip}{3pt}

\maketitle

\begin{abstract}
Synthetic data (\syn) drawn from simulators have emerged as a popular alternative for training models where acquiring annotated real-world images is difficult. However, transferring models trained on synthetic images to real-world applications can be challenging due to appearance disparities. A commonly employed solution to counter this \sr gap is unsupervised domain adaptation, where models are trained using labeled \syn data and unlabeled \real data. Mispredictions made by such \sr adapted models are often associated with miscalibration -- stemming from overconfident predictions on real data. In this paper, we introduce \augcal, a simple training-time patch for unsupervised adaptation that improves \sr adapted models by -- (1) reducing overall miscalibration, (2) reducing overconfidence in incorrect predictions and (3) improving confidence score reliability by better guiding misclassification detection -- all while retaining or improving \sr performance. Given a base \sr adaptation algorithm, at training time, \augcal involves replacing vanilla \syn images with strongly augmented views (\aug intervention) and additionally optimizing for a training time calibration loss on augmented \syn predictions (\calib intervention). We motivate \augcal using a brief analytical justification of how to reduce miscalibration on unlabeled \real data. Through our experiments, we empirically show the efficacy of \augcal across multiple adaptation methods, backbones, tasks and shifts.
\end{abstract}

\vspace{\sectionReduceTop}
\section{Introduction}
\vspace{\sectionReduceBot}
Most effective models for computer vision tasks (classification, segmentation, \etc) need to learn from a large amount of exemplar data~\citep{dosovitskiy2020image,radford2021learning,kirillov2023segment, pinto2008real} that 
captures
real-world
natural variations which may occur at deployment time. However, collecting and annotating such diverse real-world data 
can be prohibitively expensive -- for instance, densely annotating a frame of Cityscapes~\citep{cordts2016cityscapes} can take upto $\sim1.5$ hours! Machine-labeled synthetic images 
generated from
off-the-shelf simulators can substantially reduce this need for manual annotation and physical data collection~\citep{sankaranarayanan2018GTA,ros2016synthia,carlaDataset,savva2019habitat,deitke2020robothor,chattopadhyay2021robustnav}. Nonetheless, models trained on \syn data often exhibit subpar performance on \real data, primarily due to appearance discrepancies, commonly referred to as the \sr 
gap.
For instance, on GTAV (\syn) $\to$ Cityscapes (\real), an HRDA \syn-only model~\citep{hoyer2022hrda} achieves an mIoU of only $53.01$, compared to $\sim81$ mIoU attained by an equivalent model trained exclusively on \real data.

While there is significant effort in improving the realism of simulators~\citep{savva2019habitat, richter2022enhancing}, there is an equally large effort seeking to narrow this \sr performance gap by designing algorithms 
that facilitate \sr transfer.
These methods encompass both \textit{generalization}~\citep{chattopadhyay2022pasta,huang2021fsdr,zhao2022shade}. -- aiming to ensure strong out-of-the-box \real performance of \syn trained models -- and \textit{adaptation}~\citep{hoyer2022hrda,hoyer2022mic,vu2019advent,rangwani2022closer} -- attempting to adapt models using labeled \syn data and unlabeled \real data. Such generalization and adaptation methods have demonstrated notable success in reducing the \sr performance gap. For instance, \pst~\citep{chattopadhyay2022pasta} (a \textit{generalization} method) improves \sr performance of a \syn-only model from $53.01\to57.21$ mIoU. Furthermore, HRDA + MIC~\citep{hoyer2022mic} (an \textit{adaptation} approach) pushes performance even higher to $75.56$ mIoU.

\begin{figure}%
\centering
\vspace{-.5cm}
\includegraphics[width=\linewidth]{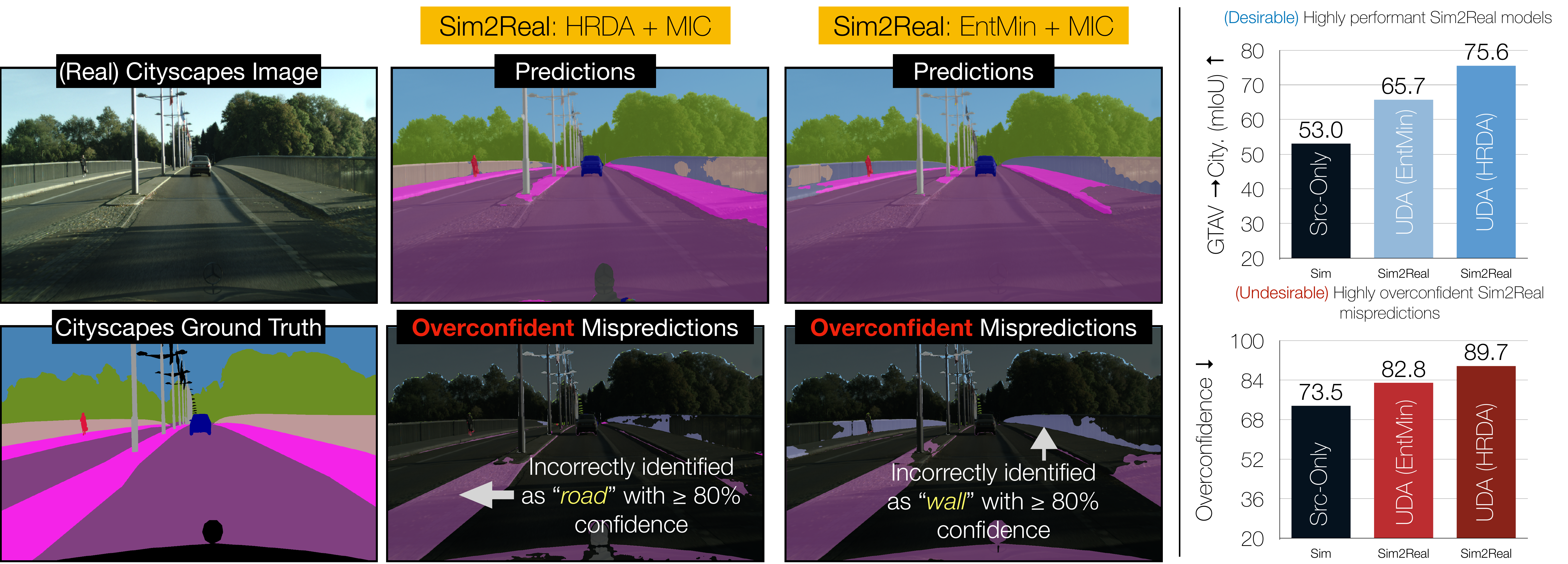}
\resizebox{0.98\textwidth}{!}{
	\begin{tabular}{@{}cccccccccc@{}}
		\cellcolor{city_color_1}\textcolor{white}{~~road~~} &
		\cellcolor{city_color_2}~~sidewalk~~&
		\cellcolor{city_color_3}\textcolor{white}{~~building~~} &
		\cellcolor{city_color_4}\textcolor{white}{~~wall~~} &
		\cellcolor{city_color_5}~~fence~~ &
		\cellcolor{city_color_6}~~pole~~ &
		\cellcolor{city_color_7}~~traffic lgt~~ &
		\cellcolor{city_color_8}~~traffic sgn~~ &
		\cellcolor{city_color_9}~~vegetation~~ & 
		\cellcolor{city_color_0}\textcolor{white}{~~ignored~~}\\
		\cellcolor{city_color_10}~~terrain~~ &
		\cellcolor{city_color_11}~~sky~~ &
		\cellcolor{city_color_12}\textcolor{white}{~~person~~} &
		\cellcolor{city_color_13}\textcolor{white}{~~rider~~} &
		\cellcolor{city_color_14}\textcolor{white}{~~car~~} &
		\cellcolor{city_color_15}\textcolor{white}{~~truck~~} &
		\cellcolor{city_color_16}\textcolor{white}{~~bus~~} &
		\cellcolor{city_color_17}\textcolor{white}{~~train~~} &
		\cellcolor{city_color_18}\textcolor{white}{~~motorcycle~~} &
		\cellcolor{city_color_19}\textcolor{white}{~~bike~~}
	\vspace{0.5mm}
	\end{tabular}
	}
\caption{\footnotesize\textbf{Overconfident \sr mispredictions}. \textbf{[Left]} We show an example of what we mean by overconfident mispredictions. For \sr adaptation on GTAV$\to$Cityscapes, we choose (DAFormer) HRDA + MIC~\citep{hoyer2022mic} and EntMin + MIC~\citep{vu2019advent} (highly performant \sr methods) and show erroneous predictions on Cityscapes (bottom row). We can see that the model identifies \textit{sidewalk} pixels as \textit{road} (2nd column) and \textit{fence} pixels as \textit{wall} (3rd column) with very high confidence. \textbf{[Right]} We show how pervasive this ``overconfidence'' phenomena is. While better \sr adapted models -- from (DAFormer) Source-Only~\citep{hoyer2022hrda} to (DAFormer) EntMin + MIC~\citep{vu2019advent} to (DAFormer) HRDA + MIC~\citep{hoyer2022mic} -- exhibit improved transfer performance \textbf{[Top, Right]}, they also exhibit increased overconfidence in mispredictions \textbf{[Bottom, Right]}, affecting prediction reliability.
}
\label{fig:problem_teaser}
\vspace{-20pt}
\end{figure}

While \sr performance may increase both from generalization or adaptation methods,
for safety-critical deployment scenarios, task performance is often not the sole factor of interest. It is additionally important to ensure \sr adapted models make \textit{calibrated} and \textit{reliable} predictions on \real data. Optimal calibration on real data ensures that the model's 
confidence in its predictions aligns with the true likelihood of correctness.  
Deploying poorly calibrated models can have severe consequences, especially in high-stakes applications
(such as autonomous driving), where users can place trust in (potentially) unreliable predictions~\citep{theguardianTeslaDriver,michelmore2018evaluating}. We find that mistakes made by \sr adaptation methods are often associated with miscalibration caused by \textit{overconfidence} -- highly confident incorrect predictions (see Fig.~\ref{fig:problem_teaser} Left). More interestingly, we find that as adaptation methods improve in terms \sr performance, the propensity to make overconfident mispredictions also increases (see Fig.~\ref{fig:problem_teaser} Right). Our focus in this paper is to devise training time solutions to mitigate this issue.

Calibrating deep neural networks (for such \sr adaptation methods) is crucial, as they routinely make overconfident predictions~\citep{guo2017calibration,gawlikowski2021survey,minderer2021revisiting}. 
While various techniques address miscalibration on ``labeled data splits'' for in-distribution scenarios, maintaining calibration in the face of dataset shifts, like \sr, proves challenging due to lack of labeled examples in the target (\real) domain. 
To address this, we propose \augcal, a training-time patch to ensure existing \sr adaptation
methods 
make \textit{accurate}, \textit{calibrated} and \textit{reliable} predictions on real data. When applied to a \sr adaptation framework, \augcal aims to satisfy three key criteria: (1) retain performance of the base \sr method, (2) reduce miscalibration and overconfidence and (3) ensure calibrated confidence scores translate to improved reliability. 
Additionally, to ensure broad applicability, \augcal aims to do so by making 
two
minimally
invasive changes to a \sr adaptation training pipeline.
First, by \textsc{Aug}menting~\citep{cubuk2020randaugment,chattopadhyay2022pasta} input \syn images during training using an \aug transform that reduces distributional distance between \syn and \real images. Second, by additionally optimizing for a \textsc{Cal}ibration loss~\citep{hebbalaguppe2022stitch,liang2020neural,liu2022devil} at training time on \textsc{Aug}mented \syn predictions.
We devise \augcal based on an analytical rationale (see Sec.~\ref{sec:real_miscalib} and~\ref{sec:why_augcal}) illustrating how it helps reduce an upper bound on desired target (\real) calibration error. Through our experiments on GTAV$\to$Cityscapes and VisDA \sr, 
we demonstrate how \augcal helps reduce miscalibration on \real data.
To summarize, we make the following contributions:
\begin{packed_itemize}
\item We propose \augcal, a training time patch, compatible with existing \sr adaptation methods that ensures \sr adapted models make \textit{accurate} (measured via adaptation performance), \textit{calibrated} (measured via calibration error) and \textit{reliable} (measured via confidence guided misclassification detection) predictions.
\item We conduct \sr adaptation experiments for object recognition (VisDA~\citep{peng2017visda}) and semantic segmentation (GTAV~\citep{sankaranarayanan2018GTA}$\to$Cityscapes~\citep{cordts2016cityscapes}) with three representative UDA methods (pseudo-label based self-training, entropy minimization and domain adversarial training) and show that applying \augcal -- (1) improves or preserves adaptation performance, (2) reduces miscalibration  and overconfidence and (3) improves the reliability of confidence scores.
\item We show how \augcal improvements are effective across multiple backbones, \aug and \calib options 
and highlight choices that are more consistently effective across experimental settings.
\end{packed_itemize}

\vspace{\sectionReduceTop}
\section{Related work}
\vspace{\sectionReduceBot}
\par \noindent
\textbf{Unsupervised Domain Adaptation (UDA).} 
We focus on UDA algorithms to address \textit{covariate shifts} in the \sr context~\citep{chattopadhyay2022pasta,choi2021robustnet,zhao2022shade,huang2021fsdr,rangwani2022closer,hoyer2022mic,sankaranarayanan2018GTA,ros2016synthia}. 
This involves adapting a model to an unseen target (\real) domain using labeled samples from a source (\syn) domain and unlabeled samples from the target domain.
Here, the source and target datasets share the same label space and labeling functions, but differences exist in the distribution of inputs~\citep{farahani2021brief,zhang2019bridging}.
\sr UDA methods~\citep{ganin2014unsupervised,hoffman2017cycada,saenko2010adapting,tzeng2014deep} range from \textit{feature distribution matching}~\citep{ganin2014unsupervised,long2018conditional,saito2018adversarial,tzeng2017adversarial,zhang2019bridging}, explicitly addressing \textit{domain discrepancy}~\citep{kang2019contrastive,long2015learning,tzeng2014deep,rangwani2022closer}, \textit{entropy minimization}~\citep{vu2019advent} or \textit{pseudo-label guided self-training}~\citep{hoyer2022hrda,hoyer2022daformer,hoyer2022mic}. We 
observe that existing \sr UDA methods usually improve performance at the expense of increasingly overconfident mispredictions on (\real) target data~\citep{wang2020transferable}. Our proposed method, \augcal, is designed to retain \sr adaptation performance while reducing miscalibration on real data for existing methods. We conduct experiments on three representative UDA methods -- Entropy Minimization~\citep{vu2019advent}, Self-training~\citep{hoyer2022hrda} and Domain Adversarial Training~\citep{rangwani2022closer}.

\par \noindent
\textbf{Confidence Calibration for Deep Networks.} 
For discriminative models, confidence calibration indicates the degree to which confidence scores associated with predictions align with the true likelihood of correctness (usually measured via ECE~\citep{naeini2015obtaining}).
Deep networks tend to be be very poor at providing calibrated confidence estimates (are overconfident) for their predictions~\citep{guo2017calibration,gawlikowski2021survey,minderer2021revisiting}, which in turn leads to less reliable predictions for decision-making in safety-critical settings. Recent work~\citep{guo2017calibration} has also shown that calibration worsens for larger models and can decrease with increasing performance. Several works~\citep{guo2017calibration,DBLP:conf/nips/Lakshminarayanan17,DBLP:conf/nips/MalininG18} have explored this problem for modern architectures, and several solutions have also been proposed --including temperature scaling (prediction logits being divided by a scalar learned on a held-out set~\citep{platt1999probabilistic,kull2017beta,bohdal2021meta,islam2021class}) and trainable calibration objectives (training time loss functions that factor in calibration ~\citep{liang2020neural,karandikar2021soft}). 
Improving network calibration is even more challenging in out-of-distribution settings due to the simultaneous lack of ground truth labels and overconfidence on unseen samples~\citep{wang2020transferable}. 
Specifically, instead of methods that rely on temperature-scaling~\citep{wang2020meta,wang2022calibrating} or maybe require an additional calibration split, \augcal explores the use of training time calibration objectives~\citep{munir2022towards} to reduce micalibration for \sr shifts.

\vspace{\sectionReduceTop}
\section{Method}
\vspace{\sectionReduceBot}
\subsection{Background}
\par \noindent
\textbf{Notations.} Let 
$x$
denote input images
and $y$ denote 
corresponding labels (from the label space $\mathcal{Y} = \{1, 2, ..., K\}$) drawn from a joint distribution $P(x, y)$.
We focus on the classification case, where the goal is to learn a discriminative model $\mathcal{M}_{\theta}$ (with parameters $\theta$) that maps input images to the desired $K$ output labels, $\mathcal{M}_{\theta}: \mathcal{X} \rightarrow \mathcal{Y}$, using a softmax layer on top.
The predictive probabilities for the given input can be expressed as $p_{\theta}(y|x) = \softmax (\mathcal{M}_{\theta}(x)$).
We use 
$\hat{y} = \argmax_{y\in\mathcal{Y}} p_{\theta}(y|x)$ to denote the predicted label for $x$ and $c$ to denote the confidence in prediction.

\par \noindent
\textbf{Unsupervised \sr Adaptation.} In unsupervised domain adaptation (UDA) for \sr settings, we assume access to a labeled (\syn) source dataset $D_S = \{(x^S_i, y^S_i) \}_{i=1}^{|S|}$ and an unlabeled (\real) target dataset $D_T = \{x^T_i\}_{i=1}^{|T|}$. We assume $D_S$ and $D_T$ splits are drawn from source and target distributions $P^S(x,y)$ and $P^T(x,y)$ respectively. At training, we have access to $D = D_S \cup D_T$. We operate in the setting where source and target share the same label space, and discrepancies exist 
only in 
input images. 
The model $\mathcal{M}_{\theta}$ is trained on labeled source images using cross entropy,
\begin{equation}
    \sum_{i=1}^{|S|} \mathcal{L}_{CE} (x_i^S, y_i^S; \theta) = - \sum_{i=1}^{|S|} y_i^S \log p_{\theta}(\hat{y}_i^S | x_i^S) \text{ where }\hat{y}_i^S = \argmax_{y\in\mathcal{Y}} p_{\theta}(y_i^S|x_i^S)
\end{equation}
UDA methods additionally optimize for an adaptation objective on labeled source and unlabeled target data ($\mathcal{L}_{UDA}$). The overall learning objective can be expressed as,
\begin{equation}
    \min_{\theta} \underbrace{\sum_{i=1}^{|S|} \mathcal{L}_{CE} (x_i^S, y_i^S; \theta)}_{\text{Source Loss}} + \underbrace{\sum_{i=1}^{|T|} \sum_{j=1}^{|S|} \lambda_{UDA} \mathcal{L}_{UDA} (x_i^T, x_j^S, y_j^S; \theta)}_{\text{Source Target Adaptation Loss}}
    \label{Eqn:sr_adapt}
\end{equation}
Different adaptation methods usually differ in terms of specific instantiations of this objective. While \augcal is applicable to any \sr adaptation method in principle, we conduct experiments with three popular methods -- Entropy Minimization~\citep{vu2019advent}, Pseudo-Label driven Self-training~\citep{hoyer2022hrda} (for semantic segmentation) and Domain Adversarial Training~\citep{rangwani2022closer} (for object recognition). We provide more details on these methods in Sec.~\ref{sec:sr_adapt} of appendix.

\par \noindent
\textbf{Uncertainty Calibration.} For a perfectly calibrated classifier, the confidence in predictions should match the empirical frequency of correctness.
Empirically, calibration can be measured using Expected Calibration Error (ECE)~\citep{naeini2015obtaining}.
To measure ECE on a test set $D = \{(x_i, y_i)\}_{i=1}^{|D|}$, we first partition the test data into $B$ bins, $D_{b} = \{ (x, y) \;|\; r_{b-1} \leq c < r_{b} \}$, using the confidence values $c$ such that $b\in\{1,...,B\}$ and $0 = r_{0} \leq r_{1} \leq r_{2} \leq \dots \leq r_{B} = 1$. Then, ECE measures the absolute differences between accuracy and confidence across instances in every bin,
\begin{equation}
    \text{ECE } =  \sum_{j=1}^{B} \frac{B}{|D|} \left \lvert \frac{1}{B} \sum_{i\in D_j}\mathbf{1}_{(y_i = \hat{y}_i)} - \sum_{i\in D_j}c_i \right \rvert
    \label{eq:ece}
\end{equation}
We are interested in models that exhibit high-performance and low calibration error (ECE). Note that Eqn~\ref{eq:ece} alone does not indicate if a model is overconfident. We define overconfidence (OC) as the expected confidence on mispredictions.
Prior work on improving calibration in out-of-distribution (OOD) settings~\citep{wang2022calibrating} and domain adaptation scenarios~\citep{wang2020transferable} typically rely on techniques like temperature scaling. These methods often necessitate additional steps, such as employing a separate calibration split or domains~\citep{gong2021confidence} or training extra models (e.g., logistic discriminators for source and target features~\citep{wang2020transferable}). In contrast, we consider using training time calibration objectives~\citep{liangimproved,hebbalaguppe2022stitch,liu2022devil} that can be optimized in addition to task-specific objectives for improved calibration.

\subsection{\augcal}
\subsubsection{Reducing Miscalibration on (Real) Target}
\label{sec:real_miscalib}
Recall that
$P^S(x,y)$ and $P^T(x,y)$ denote the source (\syn) and target (\real) data distributions. We assume $P(x, y)$ factorizes as $P(x, y) = P(x)P(y|x)$. We assume covariate shift conditions between $P^S$ and $P^T$, \ie, $P^T(x)\neq P^S(x)$ while $P^T(y|x) = P^S(y|x)$ -- discrepancies across distributions exist only in input images.
When training a model, we can only draw ``labeled samples'' $(x, y)$ from $P^S(x,y)$. We do not have access to labels from $P^T(x,y)$. Our goal is to reduce miscalibration on (unlabeled) target data using training time calibration losses. Let $\mathcal{L}_{\calib}(x, y)$ denote such a calibration loss we can minimize (on labeled data).
Using importance sampling~\citep{cortes10}, we can get an estimate of the desired calibration loss on target data as,
\begin{align}
    \E_{x,y\sim P^{T}(x,y)}\left[\mathcal{L}_{\calib}(x,y)\right]
    &= \int_x\int_y\mathcal{L}_{\calib}(x,y)P^{T}(x,y)\,dx\,dy\nonumber\\
    &=\int_x\int_y\mathcal{L}_{\calib}(x,y)\frac{P^{T}(x)P^{T}(y|x)}{P^{S}(x)P^{S}(y|x)}P^{S}(x,y)\,dx\,dy\nonumber\\
    &=\E_{x,y\sim P^{S}(x,y)}\left[\underbrace{w_S(x)}_{\text{Importance Weight}}\underbrace{\mathcal{L}_{\calib}(x,y)}_{\text{Source Loss}}\right]
    \label{eqn:loss_tgt_imp}
\end{align} 
where $w_S(x) = \frac{P^T(x)}{P^S(x)}$ denotes the importance weight. Assuming $\mathcal{L}_{\calib}(x,y) \geq 0$\footnote{We make the reasonable assumption that the calibration loss function is always non-negative.}, we can obtain an upper bound on step~\ref{eqn:loss_tgt_imp}~\citep{pampari20,wang2020transferable} as
\begin{align}
    \E_{x,y\sim P^{T}(x,y)}\left[\mathcal{L}_{\calib}(x,y)\right]
    =&\E_{x,y\sim P^{S}(x,y)}\left[w_S(x)\mathcal{L}_{\calib}(x,y)\right]\\
    \leq&\sqrt{\E_{P^{S}(x)}\Big[w_S(x)^2\Big]\E_{P^{S}(x,y)}\Big[\mathcal{L}_{\calib}(x,y)^2\Big]}\label{Eq:CS}\\
    \leq&\frac{1}{2}\Big(\underbrace{\E_{P^{S}(x)}\Big[w_S(x)^2\Big]}_{\text{Shift Dependent}}+\underbrace{\E_{P^{S}(x,y)}\Big[\mathcal{L}_{\calib}(x,y)^2\Big]}_{\text{Source Dependent}}\Big)\label{Eq:AMGM}
\end{align} 

where steps~\ref{Eq:CS} and~\ref{Eq:AMGM} use the Cauchy-Schwarz and AM-GM inequalities respectively. For a given model, the second RHS term in inequation~\ref{Eq:AMGM} is computed purely on labeled samples from the source distribution and can therefore be optimized to convergence over the course of training. The gap in $\mathcal{L}_{\calib}$ across source and target is dominated by the importance weight (first term). Following~\citep{cortes10}, the first term can also be expressed as,
\begin{equation}
    \E_{P^{S}}\big[w_S(x)^2\big]=d_2\big(P^{T}(x)||P^{S}(x)\big)
    \label{Eq:var}
\end{equation}
where $d_{\alpha}(P||Q)=\big[\sum_x\frac{P^{\alpha}(x)}{Q^{\alpha-1}(x)}\big]^{\frac{1}{\alpha-1}}$ with $\alpha>0$ is the exponential in base 2 of the Renyi-divergence~\citep{renyi60} between distributions $P$ and $Q$, i.e., $d_{\alpha}(P||Q) = 2^{D_{\alpha}(P||Q)}$, where $D_{\alpha}(P||Q) = \frac{1}{\alpha - 1} \log_2 \Sigma_x P(x) (\frac{P(x)}{Q(x)})^{\alpha - 1}$. The calibration error gap between source and target distributions is therefore, dominated by the divergence between source and target distributions. Consequently, inequation~\ref{Eq:AMGM} can be expressed as,
\begin{equation}
    \underbrace{\E_{x,y\sim P^{T}(x,y)}\left[\mathcal{L}_{\calib}(x,y)\right]}_{\text{Target Calibration Loss}} \leq \underbrace{\frac{1}{2}\textcolor{blue}{d_2\big(P^{T}(x)||P^{S}(x)\big)}}_{\text{Source and Target Divergence}} + \underbrace{\frac{1}{2}\textcolor{red}{\E_{P^{S}(x,y)}\Big[\mathcal{L}_{\calib}(x,y)^2\Big]}}_{\text{Source Calibration Loss}} = \underbrace{U (S, T)}_{\text{Upper Bound}}
    \label{Eq:calib_gap}
\end{equation}

where $U (S, T)$ denotes the upper bound on target calibration loss. Therefore, to effectively reduce miscalibration on target data, one needs to reduce the upper bound, $U(S, T)$, which translates to (1) reducing miscalibration on source data (second red term in~\ref{Eq:calib_gap}) and (2) reducing the distributional distance between input distributions across source and target (first blue term in~\ref{Eq:calib_gap}). 

\subsubsection{Why \augcal?}
\label{sec:why_augcal}
Based on the previous discussion, to improve calibration on target data, one can always invoke a training time calibration intervention (\calib) on labeled source data to reduce $\textcolor{red}{\E_{P^{S}(x,y)}[\mathcal{L}_{\calib}(x,y)]}$. In practice, after training, we can safely assume that $\textcolor{red}{\E_{P^{S}(x,y)}[\mathcal{L}_{\calib}(x,y)]} = \epsilon \rightarrow 0$ (for some very small $\epsilon$). We note that while this is useful and necessary, it is not sufficient. This is precisely where we make our contribution. To reduce both (red and blue) terms in~\ref{Eq:calib_gap}, we introduce \augcal. To do this, in addition to a training time calibration loss, $\mathcal{L}_{\calib}$, \augcal assumes that access to an additional \aug transformation that satisfies the following properties:
\begin{enumerate}
    \item $d_2\big(P^{T}(x)||P^{S}(\aug(x))\big) \leq d_2\big(P^{T}(x)||P^{S}(x)\big)$\label{prop:p1}
    \item \text{After training, }$\E_{P^{S}(x,y)}\Big[\mathcal{L}_{\calib}(x,y)\Big] \approx \E_{P^{S}(x,y)}\Big[\mathcal{L}_{\calib}(\aug(x),y)\Big] = \epsilon \rightarrow 0$\label{prop:p2}
    \label{prop:aug_prop}
\end{enumerate}

Property 1 states that the chosen \aug transformation brings transformed source data closer to target (or reduces \sr distributional distance). Property 2 states that over the course of training, irrespective of the data $\mathcal{L}_{\calib}(x, y)$ is optimized on (\aug transformed or clean source), $\E_{P^{S}(x,y)}\Big[\mathcal{L}_{\calib}(\cdot, \cdot)\Big]$ can achieve a sufficiently small value close to $0$. Given an \aug transformation that satisfies the above stated properties, we can claim,
\begin{equation}
    U^{\aug}(S, T) \leq U (S, T)
\end{equation}
where
\begin{equation}
    U^{\aug} (S, T) = \frac{1}{2}\textcolor{blue}{d_2\big(P^{T}(x)||P^{S}(\aug(x))\big)} + \frac{1}{2}\textcolor{red}{\E_{P^{S}(x,y)}\Big[\mathcal{L}_{\calib}(\aug(x),y)^2\Big]}
    \label{eqn:aug_valid}
\end{equation}

\begin{wraptable}{r}{0.5\textwidth}
\caption{\footnotesize\textbf{Eligible \aug choices.} \aug transformations that reduce \sr distance and satisfy Property-1.}\label{wrap-tab:aug_dist}
\vspace{-10pt}
\resizebox{0.5\textwidth}{!}{
\begin{tabular}{cccc}\\\toprule  
\textbf{\sr}  & \multicolumn{3}{c}{\textbf{100x (RBF) MMD}} \\
 & (\syn, \real) & (\pst-\syn, \real) & (R.Aug-\syn, \real) \\\midrule
VisDA & 4.806 & \textcolor{blue}{\textbf{4.171}} & \textcolor{blue}{\textbf{3.464}}\\
GTAV$\to$City. & 5.795 & \textcolor{blue}{\textbf{5.214}} & \textcolor{blue}{\textbf{5.213}}\\
\bottomrule
\end{tabular}}
\end{wraptable} 
That is, an appropriate \aug transform, when coupled with $\mathcal{L}_{\calib}$, helps reduce a tighter upper bound on the target calibration error than \calib. We call this intervention -- coupling \aug and \calib -- \augcal.
Naturally, the effectiveness of \augcal is directly dependent on the choice of \aug that satisfies the aforementioned properties. While most augmentations can satisfy property 2, to check if an augmentation is valid according to property 1, we compute RBF Kernel based MMD distances for (\syn, \real) and (\textsc{Aug}mented \syn, \real) feature pairs using a trained model.\footnote{Under bounded importance weight assumptions, MMD can be interpreted as an upper bound on $D_{2}(P||Q)$ and $D_{1}(P||Q)$ (KL divergence)~\citep{wang2022practical}. Note that Eqn.~\ref{eqn:aug_valid} presents upper bounds in terms of $d_{2}(P||Q) = 2^{D_{2}(P||Q)}$, and Table~\ref{wrap-tab:aug_dist} provides measured distributions distances as upper bounds on $D_{2}(P||Q)$.} In Table.~\ref{wrap-tab:aug_dist}, we show how \pst~\citep{chattopadhyay2022pasta} and RandAugment~\citep{cubuk2020randaugment}, two augmentations effective for \sr transfer, satisfy these considerations on multiple shifts (read Table.~\ref{wrap-tab:aug_dist} left to right).
\pst and RandAug are additionally (1) inexpensive when combined with \sr UDA methods, and (2) generally beneficial for \sr shifts (\pst via \sr specific design and RandAug via chained photometric operations).

\begin{figure}[t]
  \centering
\includegraphics[width=0.945\textwidth]{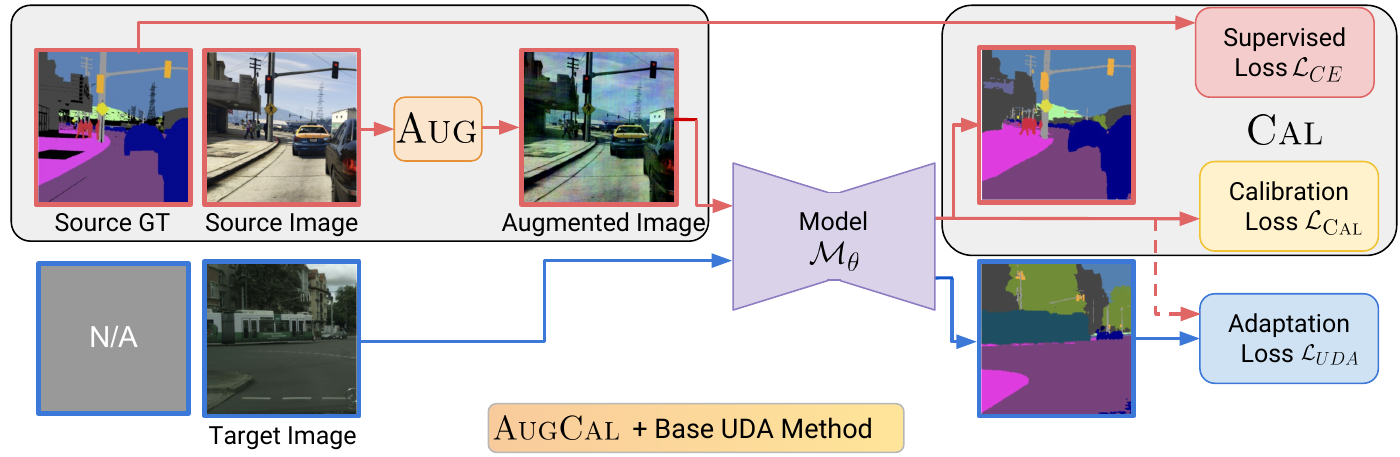}
  \caption{\footnotesize\textbf{\augcal pipeline}. \augcal consists of two key interventions on an existing \sr adaptation method. First source \syn images are augmented via an \aug transform. Supervised losses for \syn images are computed on the augmented image predictions. Additionally, \augcal optimizes for a calibration loss on \textsc{Aug}mented \syn predictions.}
  \label{fig:augcal}
  \vspace{-15pt}
\end{figure}

\vspace{\sectionReduceTop}
\subsubsection{\augcal Instantiation}
Given a \sr adaptation method, \augcal additionally optimizes for improved calibration on augmented \syn source images. Since \augcal is applicable to any existing \sr adaptation method, we abstract away the adaptation component associated with the pipeline and denote as $\mathcal{L}_{UDA}$ (see Eqn.~\ref{Eqn:sr_adapt}). The steps involved in \augcal are 
illustrated in Fig.~\ref{fig:augcal}. Given a mini-batch, we first generate augmented views, $\textsc{Aug}(x^S)$, for \syn images $x^S$. Then, during training, we optimize $\mathcal{L}_{CE}$ on those augmented \syn views and $\mathcal{L}_{UDA}$ for adaptation. To improve calibration under augmentations, we optimize an additional $\mathcal{L}_{\calib}$ loss on augmented \syn images. The overall \augcal optimization problem can be expressed as,
\begin{multline}
    \min_{\theta} \underbrace{\sum_{i=1}^{|S|} \mathcal{L}_{\tiny{CE}} (\textcolor{teal}{\aug(x_i^S)}, y_i^S; \theta)}_{\text{Source Task Loss}} + \underbrace{\sum_{i=1}^{|T|} \sum_{j=1}^{|S|} \lambda_{\tiny{UDA}} \mathcal{L}_{\tiny{UDA}} (x_i^T, \textcolor{teal}{\aug(x_j^S)}, y_j^S; \theta)}_{\text{Source Target Adaptation Loss}} \\+ \underbrace{\sum_{i=1}^{|S|}\textcolor{teal}{\lambda_{\calib}\mathcal{L}_{\tiny{\textsc{Cal}}} (\aug(x_i^S), y_i^S; \theta)}}_{\text{Source Calibration Loss}}
\end{multline}

where $\lambda_{UDA}$ and $\lambda_{\textsc{Cal}}$ denote the respective loss coefficients and the changes to a vanilla \sr adaptation framework are denoted in \textcolor{teal}{teal}. 

\par \noindent
\textbf{Choice of ``\textsc{Aug}''.} Strongly augmenting \syn images during training has proven useful for \sr transfer~\citep{chattopadhyay2022pasta,huang2021fsdr,zhao2022shade,cubuk2020randaugment}. 
For \augcal, we are interested in augmentations that satisfy the properties outlined in Sec.~\ref{sec:why_augcal}. As stated earlier, we find that RandAugment~\citep{cubuk2020randaugment} and \pst~\citep{chattopadhyay2022pasta} empirically satisfy this criteria. We use both of them for our experiments and provide details on the operations for these \aug transforms in Sec.~\ref{sec:aug_choices} of appendix.

\par \noindent
\textbf{Choice of ``\textsc{Cal}'' ($\mathcal{L}_{\textsc{cal}}$).} \augcal relies on using training time calibration losses to reduce miscalibration. Prior work in uncertainty calibration has considered several auxiliary objectives to calibrate a model being trained to reduce negative log-likelihood (NLL)~\citep{hebbalaguppe2022stitch,kumar2018trainable,karandikar2021soft,liang2020neural}. For ``\textsc{Cal}'' in \augcal, while we
consider multiple calibration losses -- DCA~\citep{liangimproved}, MbLS~\citep{liu2022devil} and MDCA~\citep{hebbalaguppe2022stitch} -- and find that DCA, simple "difference between confidence and accuracy (DCA)" loss proposed in~\citep{liang2020neural} is more consistently effective across experimental settings. DCA can be expressed as,
\begin{equation}
    \mathcal{L}_{\textsc{cal}} = \frac{1}{|S|}\left \lvert \sum_{i=1}^{|S|}\mathbf{1}_{(y^S_i = \hat{y}^S_i)} - \sum_{i=1}^{|S|}p_{\theta}(\hat{y}^S_i|x^S_i) \right \rvert
\end{equation}
where $\mathbf{1}_{(y^S_i = \hat{y}^S_i)} \text{ and } p_{\theta}(\hat{y}^S_i|x^S_i)$ denote the correctness and confidence scores associated with predictions. The DCA loss forces the mean predicted confidence over training samples to match accuracy. In the following sections, we empirically validate \augcal across adaptation methods.

\vspace{\sectionReduceTop}
\section{Experimental Details}
\label{sec:experiments}
\vspace{\sectionReduceBot}
We conduct \sr adaptation experiments across two tasks -- Semantic Segmentation (SemSeg) and Object Recognition (ObjRec). For our experiments, we train models using labeled \syn images and unlabeled \real images. We test trained models on \real images.

\par \noindent
\textbf{\sr Shifts.} For SemSeg, we conduct experiments on the GTAV$\to$Cityscapes shift. GTAV~\citep{sankaranarayanan2018GTA} consists of $\sim25$k densely annotated \syn ground-view images and Cityscapes~\citep{cordts2016cityscapes} consists of $\sim5$k \real ground view images. We report 
all metrics on the
Cityscapes validation split.
For ObjRec, we conduct experiments on the VisDA \sr benchmark. VisDA~\citep{peng2017visda} consists of $\sim152$k \syn images and $\sim55$k \real images across $12$ classes. We report all metrics on the validation split of (\real) target images. 

\par \noindent
\textbf{Models.} We check \augcal compatibility with both CNN and Transformer based 
architectures. For SemSeg, we consider DeepLabv2~\citep{chen2017deeplab} (with a ResNet-101~\citep{he2016deep} backbone) and DAFormer~\citep{hoyer2022daformer} (with an MiT-B5~\citep{xie2021segformer} backbone) architectures. For ObjRec, we consider ResNet-101 and ViT-B/16~\citep{dosovitskiy2020image} backbones with bottleneck layers as classifiers.
We start with backbones pre-trained on ImageNet~\citep{imagenet}.
\par \noindent
\textbf{Adaptation Methods.} We consider three 
representative \sr adaptation methods for our experiments. 
For SemSeg, we consider entropy minimization (EntMin)~\citep{vu2019advent} and high-resolution domain adaptive semantic segmentation (HRDA)~\citep{hoyer2022hrda}. For ObjRec, we consider smooth domain adversarial training (SDAT)~\citep{rangwani2022closer}.
For both tasks, we further improve performance with masked image consistency (MIC)~\citep{hoyer2022mic} on target images during training. 
We use MIC~\citep{hoyer2022mic}'s implementations of the adaptation algorithms and provide more training details in Sec.~\ref{sec:training_details} of appendix.
\par \noindent
\textbf{Calibration Metrics.} We use ECE to report overall confidence calibration on \real images. 
Since we are interested in reducing overconfident mispredictions, 
we also report calibration error on incorrect samples (IC-ECE)~\citep{wang2022calibrating} and mean overconfidence for mispredictions (OC).

\textbf{Reliability Metrics.} While reducing overconfidence and improving calibration on real data is desirable, this is a proxy for the true goal of improving model reliability. To assess reliability, following prior work~\citep{de2023reliability,malinin2019ensemble}, we measure whether calibrated confidence scores can better guide misclassification detection. 
To measure this, we use Prediction Rejection Ratio (PRR)~\citep{malinin2019ensemble}, which if high (positive and close to $100$) indicates that confidence scores can be used as reliable indicators of performance (details in Sec.~\ref{sec:prr} of appendix). 

Unless specified otherwise, we use \pst as the choice of \aug and DCA~\citep{liangimproved} as the choice of \calib in \augcal. We use $\lambda_{\calib} = 1$ for DCA. 

\vspace{\sectionReduceTop}
\section{Findings}
\vspace{\sectionReduceBot}
\subsection{Improving \sr Adaptation}
Recall that when applied to a \sr adaptation method, we expect \augcal to -- (1) retain \sr transfer performance, (2) reduce miscalibration and overconfidence and (3) ensure calibrated confidence scores translate to improved model reliability. We first verify these criteria.
\begin{table}
\setlength{\tabcolsep}{1pt}
\caption{\footnotesize\textbf{\augcal ensures \sr adapted models make accurate, calibrated and reliable predictions.} We find that applying \augcal to multiple \sr adaptation methods across tasks leads to better calibration (ECE, IC-ECE), reduced overconfidence (OC) and improved reliability (PRR) -- all while retaining or improving transfer performance. Highlighted rows are \augcal variants of the base methods. For \augcal, we use \pst as \aug and DCA
as \calib. $\pm$ indicates standard error.}
\vspace{-10pt}
\parbox{.5\linewidth}{
\centering
\resizebox{0.5\columnwidth}{!}{
\begin{tabular}{lccccccc}
\toprule
\multirow{2}{*}{\textbf{Method}} & \multicolumn{1}{c}{\textbf{Perf. ($\uparrow$)}} & & \multicolumn{3}{c}{\textbf{Calibration Error ($\downarrow$)}} & & \multicolumn{1}{c}{\textbf{Reliability ($\uparrow$)}}\\ 
 & mIoU & & ECE & IC-ECE & OC & & PRR \\
\midrule
\multicolumn{1}{l}{\texttt{1} EntMin + MIC} & 65.71 & & 5.34\tiny{\std{0.35}} & 77.73\tiny{\std{0.26}} & 82.83\tiny{\std{0.55}} & & 45.93\tiny{\std{0.54}}\\
\rowcolor{defaultcolor}
\multicolumn{1}{l}{\texttt{2} \quad + \augcal} & \textbf{70.31} & & \textbf{3.43}\tiny{\std{0.29}} & \textbf{72.97}\tiny{\std{0.26}} & \textbf{82.80}\tiny{\std{0.57}} & & \textbf{62.66}\tiny{\std{0.55}}\\
\multicolumn{1}{l}{\texttt{3} HRDA + MIC} & 75.56 & & 2.86\tiny{\std{0.10}} & 81.92\tiny{\std{0.14}} & 89.72\tiny{\std{0.48}} & & 68.91\tiny{\std{0.46}}\\
\rowcolor{defaultcolor}
\multicolumn{1}{l}{\texttt{4} \quad + \augcal}  & \textbf{75.90} & & \textbf{2.45}\tiny{\std{0.09}} & \textbf{79.09}\tiny{\std{0.16}} & \textbf{88.26}\tiny{\std{0.49}} & & \textbf{70.35}\tiny{\std{0.51}}\\
\bottomrule
\end{tabular}}
\caption*{(a) GTAV$\to$Cityscapes. (DAFormer).}
}
\hfill
\parbox{.5\linewidth}{
\centering
\resizebox{0.5\columnwidth}{!}{
\begin{tabular}{lccccccc}
\toprule
\multirow{2}{*}{\textbf{Method}} & \multicolumn{1}{c}{\textbf{Perf. ($\uparrow$)}} & & \multicolumn{3}{c}{\textbf{Calibration Error ($\downarrow$)}} & & \multicolumn{1}{c}{\textbf{Reliability ($\uparrow$)}}\\ 
 & mIoU & & ECE & IC-ECE & OC & & PRR \\
\midrule
\multicolumn{1}{l}{\texttt{1} SDAT + MIC} & 92.53\tiny{\std{0.28}} & & 7.67\tiny{\std{0.49}} & 91.45\tiny{\std{0.63}} & 89.13\tiny{\std{1.29}} & & 63.78\tiny{\std{2.12}}\\
\rowcolor{defaultcolor}
\multicolumn{1}{l}{\texttt{2} \quad + \augcal} & \textbf{92.87}\tiny{\std{0.06}} & & \textbf{6.84}\tiny{\std{0.10}} & \textbf{89.25}\tiny{\std{0.36}} & \textbf{85.74}\tiny{\std{0.36}} & & \textbf{67.80}\tiny{\std{0.78}}\\
\bottomrule
\end{tabular}}
\caption*{(b) VisDA \sr. (ViT-B).}
}
\label{table:aug_cal_key_res}
\vspace{-20pt}
\end{table}

\par \noindent
\textbf{$\triangleright$ \augcal improves or retains \sr adaptation performance.} Since \augcal intervenes on an existing \sr adaptation algorithm, we first verify that encouraging better calibration does not adversely impact \sr adapation performance. We find that performance is either retained or improved (\eg, for EntMin + MIC in Table.~\ref{table:aug_cal_key_res} (a)) as miscalibration is reduced (Tables.~\ref{table:aug_cal_key_res}(a) and (b), Perf. columns). 

\par \noindent
\textbf{$\triangleright$ \augcal reduces miscalibration post \sr adaptation.} On both GTAV$\to$Cityscapes and VisDA, we find that \augcal consistently reduces miscalibration of the base method by reducing overconfidence on incorrect predictions. This is evident in how \augcal variants of the base adaptation methods have lower ECE, IC-ECE and OC values (\augcal rows, Calibration columns in Tables.~\ref{table:aug_cal_key_res} (a) and (b)). As an example, 
to illustrate the effect of improved calibration on real data, in Fig.~\ref{fig:augcal_qual}, we show how applying \augcal can improve the proportion of per-pixel SemSeg predictions that are accurate and have high-confidence ($>0.95$).

\par \noindent
\textbf{$\triangleright$ \augcal improvements in calibration improve reliability.} As noted earlier, we additionally investigate the extent to which calibration improvements for \sr adaptation translate to reliable confidence scores -- via misclassification detection on \real target data (see Sec.~\ref{sec:experiments}), as measured by PRR. We find that \augcal consistently improves PRR of the base \sr adaptation method (PRR columns for \augcal rows in Tables.~\ref{table:aug_cal_key_res}(a) and (b)) -- ensuring that predictions made \augcal variants of a base model are more trustworthy.

\subsection{Analyzing \augcal}
We now analyze different aspects of \augcal.
\begin{table}
\setlength{\tabcolsep}{2pt}
\caption{\footnotesize\textbf{\augcal is better than applying \textsc{Aug} or \textsc{Cal} alone.} On GTAV$\to$Cityscapes and VisDA, we show that \augcal improves over just augmented \syn training (\aug) or just optimizing for calibration on
\syn data (\calib). For \augcal, we use \pst as \aug and DCA
as \calib. $\pm$ indicates standard error.}
\vspace{-10pt}
\parbox{.5\linewidth}{
\centering
\resizebox{\linewidth}{!}{
\begin{tabular}{lccccc}
\toprule
\multirow{2}{*}{\textbf{Method}} & \multicolumn{1}{c}{\textbf{Perf. ($\uparrow$)}} & \multicolumn{2}{c}{\textbf{Calibration Error ($\downarrow$)}} & \multicolumn{1}{c}{\textbf{Reliability ($\uparrow$)}}\\
& mIoU & ECE & IC-ECE & PRR\\
\midrule
\texttt{1} EntMin + MIC & 65.71 & 5.34\tiny{\std{0.35}} & 77.73\tiny{\std{0.26}} & 45.93\tiny{\std{0.54}}\\
\texttt{2} \quad + \textsc{Aug} & 67.58 & 4.30\tiny{\std{0.33}} & 77.59\tiny{\std{0.25}} & 48.05\tiny{\std{0.53}} \\
\texttt{3} \quad + \textsc{Cal} & 68.70 & 4.04\tiny{\std{0.26}} & 75.86\tiny{\std{0.26}} & 52.52\tiny{\std{0.54}}\\
\rowcolor{defaultcolor}
\texttt{4} \quad + \augcal & \textbf{70.31} & \textbf{3.43}\tiny{\std{0.29}} & \textbf{72.97}\tiny{\std{0.26}} & \textbf{62.66}\tiny{\std{0.55}}\\
\bottomrule
\end{tabular}}
\caption*{(a) GTAV$\to$Cityscapes. (DAFormer).}
}
\hfill
\parbox{.5\linewidth}{
\centering
\resizebox{\linewidth}{!}{
\begin{tabular}{lccccc}
\toprule
\multirow{2}{*}{\textbf{Method}} & \multicolumn{1}{c}{\textbf{Perf. ($\uparrow$)}} & \multicolumn{2}{c}{\textbf{Calibration Error ($\downarrow$)}} & \multicolumn{1}{c}{\textbf{Reliability ($\uparrow$)}}\\
& mAcc & ECE & IC-ECE & PRR\\
\midrule
\texttt{1} SDAT + MIC & 92.53\tiny{\std{0.28}} & 7.67\tiny{\std{0.49}} & 91.45\tiny{\std{0.63}} & 63.78\tiny{\std{2.12}}\\
\texttt{2} \quad + \textsc{Aug} & 92.69\tiny{\std{0.15}} & 9.48\tiny{\std{1.99}} & 90.48\tiny{\std{0.33}} & 65.68\tiny{\std{0.58}}\\
\texttt{3} \quad + \textsc{Cal} & 91.63\tiny{\std{0.71}} & 7.30\tiny{\std{0.09}} & 91.19\tiny{\std{0.13}} & 66.62\tiny{\std{1.70}}\\
\rowcolor{defaultcolor}
\texttt{4} \quad + \augcal & \textbf{92.87}\tiny{\std{0.06}}& \textbf{6.84}\tiny{\std{0.10}} & \textbf{89.25}\tiny{\std{0.36}} & \textbf{67.80}\tiny{\std{0.78}}\\
\bottomrule
\end{tabular}}
\caption*{(b) VisDA \sr (ViT-B).}
}
\label{table:aug_cal_augcal}
\vspace{-25pt}
\end{table}

\par \noindent
\textbf{$\triangleright$ Applying \augcal is better than applying just \textsc{Aug} or \textsc{Cal}.} In Sec.~\ref{sec:real_miscalib} and~\ref{sec:why_augcal}, we discuss how \augcal can be more effective in reducing target miscalibration than just optimizing for improved calibration on labeled \syn images. We verify this empirically in Tables.~\ref{table:aug_cal_augcal}(a) and (b) for SemSeg and ObjRec. We show that while \aug and \calib, when applied individually, improve over a base \sr method, they fall short of improvements offered by \augcal.

\par \noindent
\textbf{$\triangleright$ \augcal is applicable across multiple \aug choices.} In Sec.~\ref{sec:why_augcal} and Table.~\ref{wrap-tab:aug_dist}, we show how both \pst~\citep{chattopadhyay2022pasta} and RandAugment~\citep{cubuk2020randaugment} are eligible for \augcal. In Table.~\ref{table:augcal_ablations}(a), we fix DCA as \calib and find that both \pst and RandAugment are effective in retaining or improving performance, reducing miscalibration and improving reliability.

\begin{table}
\setlength{\tabcolsep}{2pt}
\caption{\footnotesize\textbf{Ablating \aug and \calib choices in \augcal.} For a DAFormer model on GTAV$\to$Cityscapes, \augcal successfully reduces miscalibration and produces reliable confidence scores for \sr adaptation using both \pst (P) and RandAug (R) as \aug choices. We also ablate the choice of \calib in \augcal across DCA, MDCA and MbLS and find that DCA is more consistently effective in reducing miscalibration across tasks and settings. $\lambda_{\calib} = 1$ for MDCA and $\lambda_{\calib} = 0.1, m=10$ for MbLS. $\pm$ indicates standard error.}
\vspace{-10pt}
\parbox{.5\linewidth}{
\centering
\resizebox{\linewidth}{!}{
\begin{tabular}{lccccccc}
\toprule
\multirow{2}{*}{\textbf{Method}} & \multirow{2}{*}{\textbf{\aug}} &  \multicolumn{1}{c}{\textbf{Perf. ($\uparrow$)}} & & \multicolumn{2}{c}{\textbf{Calibration Error ($\downarrow$)}} & & \multicolumn{1}{c}{\textbf{Reliability ($\uparrow$)}}\\ 
 & & mIoU & & ECE & IC-ECE & & PRR \\
\midrule
\multicolumn{1}{l}{\texttt{1} EntMin} &&  65.71 & & 5.34\tiny{\std{0.35}} & 77.73\tiny{\std{0.26}} & & 45.93\tiny{\std{0.54}}\\
\rowcolor{defaultcolor}
\multicolumn{1}{l}{\texttt{2} \quad + \augcal} & P &  70.31 & & 3.43\tiny{\std{0.29}} & 72.97\tiny{\std{0.26}} & & 62.66\tiny{\std{0.55}}\\
\rowcolor{defaultcolor}
\multicolumn{1}{l}{\texttt{3} \quad + \augcal} & R &  70.65 & & 2.34\tiny{\std{0.14}} & 73.77\tiny{\std{0.21}} & & 66.65\tiny{\std{0.46}}\\
\midrule
\multicolumn{1}{l}{\texttt{4} HRDA} &&  75.56 & & 2.86\tiny{\std{0.10}} & 81.92\tiny{\std{0.14}} & & 68.91\tiny{\std{0.46}}\\
\rowcolor{defaultcolor}
\multicolumn{1}{l}{\texttt{5} \quad + \augcal}  & P &  75.90 & & 2.45\tiny{\std{0.09}} & 79.09\tiny{\std{0.16}} & & 70.35\tiny{\std{0.51}}\\
\rowcolor{defaultcolor}
\multicolumn{1}{l}{\texttt{6} \quad + \augcal}  & R & 74.10 & & 2.77\tiny{\std{0.17}} & 77.94\tiny{\std{0.18}} & & 69.46\tiny{\std{0.46}}\\
\bottomrule
\end{tabular}}
\caption*{(a) Ablating \aug in \augcal. (\calib = DCA).}
}
\hfill
\parbox{.5\linewidth}{
\centering
\resizebox{\linewidth}{!}{
\begin{tabular}{lccccccc}
\toprule
\multirow{2}{*}{\textbf{Method}} & \multirow{2}{*}{\textbf{\calib}} &  \multicolumn{1}{c}{\textbf{Perf. ($\uparrow$)}} & & \multicolumn{2}{c}{\textbf{Calibration Error ($\downarrow$)}} & & \multicolumn{1}{c}{\textbf{Reliability ($\uparrow$)}}\\ 
 & & mIoU & & ECE & IC-ECE & & PRR \\
\midrule
\multicolumn{1}{l}{\texttt{1} EntMin + MIC} &&  65.71 & & 5.34\tiny{\std{0.35}} & 77.73\tiny{\std{0.26}} & & 45.93\tiny{\std{0.54}}\\
\rowcolor{defaultcolor}
\multicolumn{1}{l}{\texttt{2} \quad + \augcal} & DCA &  70.31 & & 3.43\tiny{\std{0.29}} & 72.97\tiny{\std{0.26}} & & 62.66\tiny{\std{0.55}}\\
\rowcolor{defaultcolor}
\multicolumn{1}{l}{\texttt{3} \quad + \augcal} & MDCA &  69.50 & & 3.22\tiny{\std{0.26}} & 72.65\tiny{\std{0.25}} & & 59.96\tiny{\std{0.51}}\\
\rowcolor{defaultcolor}
\multicolumn{1}{l}{\texttt{4} \quad + \augcal} & MbLS &  68.77 & & 2.90\tiny{\std{0.24}} & 72.53\tiny{\std{0.23}} & & 61.57\tiny{\std{0.48}}\\
\midrule
\multicolumn{1}{l}{\texttt{5} HRDA + MIC} &&  75.56 & & 2.86\tiny{\std{0.10}} & 81.92\tiny{\std{0.14}} & & 68.91\tiny{\std{0.46}}\\
\rowcolor{defaultcolor}
\multicolumn{1}{l}{\texttt{6} \quad + \augcal}  & DCA &  75.90 & & 2.45\tiny{\std{0.09}} & 79.09\tiny{\std{0.16}} & & 70.35\tiny{\std{0.51}}\\
\multicolumn{1}{l}{\texttt{7} \quad + \augcal}  & MDCA & 75.50 & & 2.92\tiny{\std{0.15}} & 80.76\tiny{\std{0.16}} & & 68.45\tiny{\std{0.47}}\\
\multicolumn{1}{l}{\texttt{8} \quad + \augcal}  & MbLS & 71.15 & & 2.68\tiny{\std{0.15}} & 70.21\tiny{\std{0.42}} & & 69.33\tiny{\std{0.47}}\\
\bottomrule
\end{tabular}}
\caption*{(b) Ablating \calib in \augcal. (\aug = \pst)}
}
\label{table:augcal_ablations}
\vspace{-20pt}
\end{table}
\par \noindent
\textbf{$\triangleright$ Ablating \calib choices for \augcal.} For completeness, we also conduct experiments by fixing \pst as \aug and ablating the choice of \calib in \augcal. We consider recently proposed training time calibration objectives -- Difference of Confidence and Accuracy (DCA)~\citep{liangimproved}, Multi-class Difference in Confidence and Accuracy (MDCA)~\citep{hebbalaguppe2022stitch} and Margin-based Label Smoothing (MbLS)~\citep{liu2022devil} -- as potential \calib choices (results for SemSeg outlined in Table.~\ref{table:augcal_ablations}(b)). We find that while MDCA and MbLS can be helpful, DCA is more consistently helpful across tasks and settings.

\par \noindent
\textbf{$\triangleright$ \augcal is applicable across multiple task backbones.} Different architectures -- CNNs and Transformers -- are known to exhibit bias towards different properties in images (shape, texture, \etc)~\citep{naseer2021intriguing}. Since the choice of \aug transform (which can alter such properties) is central to the efficacy of \augcal, we verify if \augcal is effective across both CNN and Transformer backbones. To do this, we conduct our SemSeg, ObjRec experiments with both transformer (DAFormer, ViT-B) and CNN (DeepLabv2-R101, ResNet-101) architectures. We find that \augcal (with \pst as \aug and DCA as \calib) is effective in reducing \sr miscalibration across all settings. We discuss these results in Sec.~\ref{sec:augcal_res} of appendix.

\begin{figure}[t]
  \centering
\includegraphics[width=0.95\textwidth]{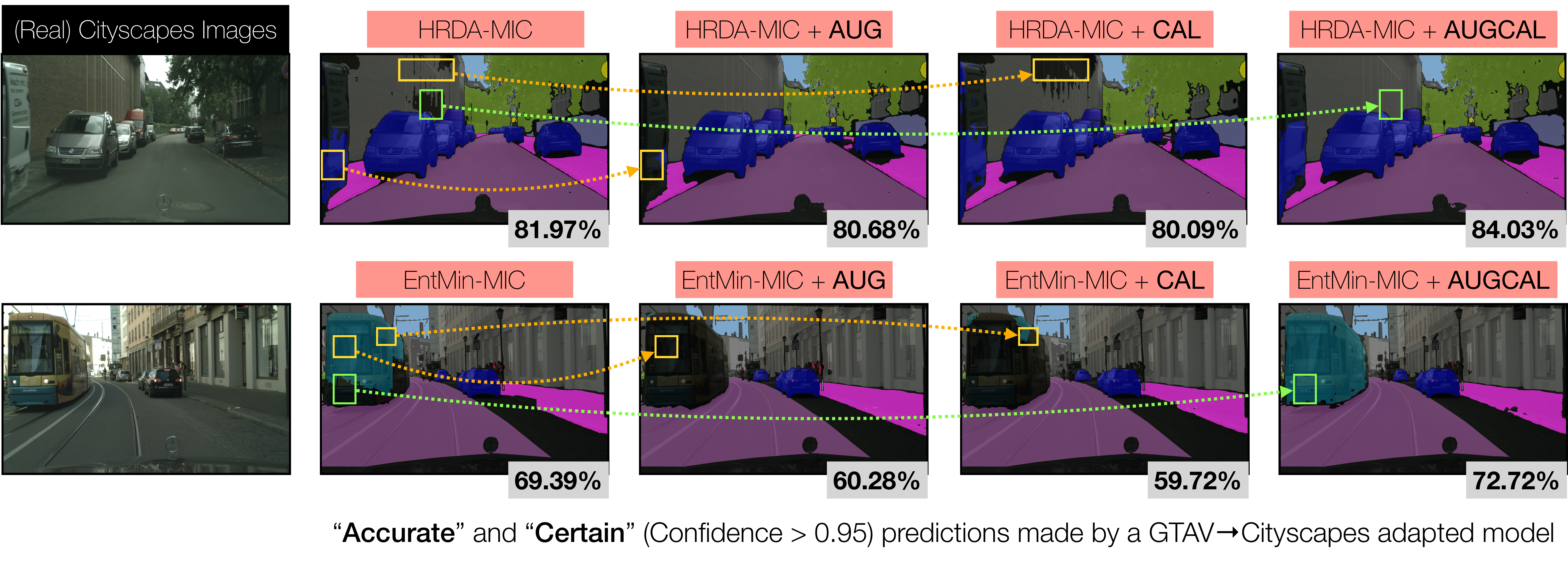}
\resizebox{0.98\textwidth}{!}{
	\begin{tabular}{@{}cccccccccc@{}}
		\cellcolor{city_color_1}\textcolor{white}{~~road~~} &
		\cellcolor{city_color_2}~~sidewalk~~&
		\cellcolor{city_color_3}\textcolor{white}{~~building~~} &
		\cellcolor{city_color_4}\textcolor{white}{~~wall~~} &
		\cellcolor{city_color_5}~~fence~~ &
		\cellcolor{city_color_6}~~pole~~ &
		\cellcolor{city_color_7}~~traffic lgt~~ &
		\cellcolor{city_color_8}~~traffic sgn~~ &
		\cellcolor{city_color_9}~~vegetation~~ & 
		\cellcolor{city_color_0}\textcolor{white}{~~ignored~~}\\
		\cellcolor{city_color_10}~~terrain~~ &
		\cellcolor{city_color_11}~~sky~~ &
		\cellcolor{city_color_12}\textcolor{white}{~~person~~} &
		\cellcolor{city_color_13}\textcolor{white}{~~rider~~} &
		\cellcolor{city_color_14}\textcolor{white}{~~car~~} &
		\cellcolor{city_color_15}\textcolor{white}{~~truck~~} &
		\cellcolor{city_color_16}\textcolor{white}{~~bus~~} &
		\cellcolor{city_color_17}\textcolor{white}{~~train~~} &
		\cellcolor{city_color_18}\textcolor{white}{~~motorcycle~~} &
		\cellcolor{city_color_19}\textcolor{white}{~~bike~~}
	\vspace{0.5mm}
	\end{tabular}
	}
  \caption{\footnotesize\textbf{\augcal increases the proportion of ``accurate'' and ``certain'' predictions}. For a (DAFormer) HRDA + MIC (row 1) and EntMin + MIC (row 2) on GTAV$\to$Cityscapes, we show how different interventions affect the proportion of ``accurate'' and ``certain'' (confidence $> 0.95$) predictions (indicated in \textcolor{gray}{gray} per column). Regions in black do not satisfy the ``accurate'' and ``certain'' filtering criteria. We see that compared to a base adaptation method, \augcal increases the proportion highly-confident correct predictions (\textcolor{green}{green} boxes). \aug and \calib applied alone can potentially reduce that proportion (\textcolor{yellow}{yellow} boxes). \aug is \pst, \calib is DCA.}
  \label{fig:augcal_qual}
  \vspace{-20pt}
\end{figure}

\par \noindent
\textbf{$\triangleright$ How does \augcal compare with temperature scaling?} While we focus on ``training-time'' patches to improve \sr calibration, we also conduct an experiment to compare \augcal with ``post-hoc'' temperature scaling (TS) on VisDA.
Specifically, we use 80\% of VisDA \syn images for training models and rest (20\%) for validation and temperature tuning. To ensure a fair comparison, we consider temperature tuning on both "clean" (C) and "\pst augmented" (P) val splits.
We find that irrespective of tuning on C or P, unlike \augcal, TS is ineffective and increases overconfidence and miscalibration. We present these results in Sec.~\ref{sec:augcal_res} of appendix.

\par \noindent
\textbf{$\triangleright$ \augcal increases the proportion of accurate and certain predictions.} In Fig.~\ref{fig:augcal_qual}, we show qualitatively for GTAV$\to$Cityscapes SemSeg how \augcal increases the proportion of highly-confident correct predictions. In practice, we find that this improvement is much more subtle for stronger \sr adaptation methods, such as HRDA + MIC, compared to weaker ones, such as EntMin + MIC, which have considerable room for improvement.

\vspace{\sectionReduceTop}
\section{Conclusion}
\vspace{\sectionReduceBot}
We propose \augcal, a method to reduce the miscalibration of \sr adapted models, often caused due to highly-confident incorrect predictions.
\augcal modifies a \sr adaptation framework by making two minimally invasive changes -- (1) augmenting \syn images via \aug transformations that reduce \sr distance and (2) optimizing for an additional calibration loss on \textsc{Aug}mented \syn predictions. Applying \augcal to existing adaptation methods for semantic segmentation and object recognition reduces miscalibration, overconfidence and improves reliability of confidence scores, all while retaining or improving performance on \real data. \augcal is meant to be a task-agnostic, general purpose framework to reduce miscalibration for \sr adaptation methods and we hope such simple methods are taken into consideration for experimental settings beyond the ones considered in this paper.
\par \noindent
\textbf{Acknowledgements.} This work has been partially sponsored by NASA University Leadership Initiative (ULI) \#80NSSC20M0161, ARL and NSF \#2144194.

\section{Ethics Statement}
Our proposed patch, \augcal, is meant to improve the reliability of \sr adapted models. We assess reliability in terms of confidence calibration (prediction scores aligning with true likelihood of correctness) and the extent to which calibrated confidence scores are useful for assessing prediction quality (measured via mis-classification detection).  \augcal adapted models have promising consequences for downstream applications. A well-calibrated and reliable \sr adapted model can increase
transparency in \real predictions and facilitate robust decision making in safety-critical scenarios. 
That said, we would like to note that while \augcal is helpful for our specific measures of reliability, exploration along other domain specific notions of reliability remain.

\bibliography{iclr2024_conference,main}
\bibliographystyle{iclr2024_conference}

\clearpage
\appendix
\section{Appendix}
\subsection{Overview}
This appendix is organized as follows. In Sec.~\ref{sec:aug_choices}, we provide more details on the \aug choices for \augcal -- \pst and RandAugment. Sec.~\ref{sec:training_details} outlines training, implementation details and choice of hyperparameters. Then, in Sec.~\ref{sec:prr}, we discuss Prediction Rejection Ratio (PRR), the metric used to assess reliability in our experiments. 
In Sec.~\ref{sec:augcal_res}, we provide \augcal results for CNN backbones, comparisons with temperature scaling and discuss sensitivity to $\lambda_{\calib}$ (coefficient of $\mathcal{L}_{\calib}$).
Then, we
provide more supporting qualitative examples in Sec.~\ref{sec:qual_pred}. In Sec.~\ref{sec:sr_adapt}, we describe the \sr adaptation methods we use in detail. Finally, Sec.~\ref{sec:assets} summarizes the assets used for our experiments and their associated licenses.
\label{sec:overview}

\subsection{\aug Choices}
\subsubsection{\pst}
We use Proportional Amplitude Spectrum Training Augmentation (\pst)~\citep{chattopadhyay2022pasta} as one of the \aug choices in \augcal. \pst applies structured perturbations (controlled by hyper-parameters $\alpha, \beta, k$) to the amplitude spectra of synthetic images to generate augmented views. For a single-channel image, $x\in\mathbb{R}^{H\times W}$ (for illustration purposes), the augmentation process in \pst is outlined below. We refer the reader to~\citep{chattopadhyay2022pasta} for more details.

\begin{enumerate}
    \item Set $\alpha = 3.0, \beta = 0.25, k=2$ for \pst
    \item Use FFT~\citep{nussbaumer1981fast} to obtain the Fourier spectrum of synthetic image $x$, as $\mathcal{F}(x) = \text{FFT}(x)\in \mathbb{C}^{H\times W}$
    \item Obtain the corresponding amplitude and phase spectra as $\mathcal{A}(x) = \text{Abs}(\mathcal{F}(x))$ and $\mathcal{P}(x) = \text{Ang}(\mathcal{F}(x))$ respectively.
    \item Zero-center the amplitude spectrum, as $\mathcal{A}(x) = \text{FFTShift} (\mathcal{A}(x))$, to get the lowest frequency components at the center.
    \item Define perturbation strength, $\sigma \in \mathbb{R}^{H\times W}$, as $\sigma[m,n] = \left(2\alpha\sqrt{\frac{m^2 + n^2}{H^2 + W^2}}\right)^k + \beta$, where $m,n$ denote the spatial frequencies.
    \item Sample perturbations using the perturbation strength as, $\epsilon \sim \mathcal{N}(1, \sigma^2)$
    \item Perturb the amplitude spectrum $\hat{\mathcal{A}}(x) = \epsilon \odot \mathcal{A}(x)$
    \item Reset the low-frequency components, $\hat{\mathcal{A}}(x) = \text{FFTShift} (\hat{\mathcal{A}}(x))$
    \item Obtain the augmented synthetic image via inverse FFT as, $\hat{x} = \text{iFFT}(\hat{\mathcal{A}}(x), \mathcal{P}(x))$
\end{enumerate}

We set \pst hyper-parameters as $\alpha = 3.0, \beta = 0.25, k = 2$ as it seems to work well across multiple \sr shifts in practice. We use Pytorch~\citep{NEURIPS2019_9015} to vectorize all operations in the steps above and apply \pst on minibatches, instead of individual synthetic images. 

\subsubsection{RandAugment}

RandAugment~\citep{cubuk2020randaugment} generates augmented views by applying a series of transforms sampled from a predefined vocabulary. While the vocabulary includes geometric as well as photometric transforms, we only use photometric transforms for our experiments. For SemSeg, the models we train already use geometric transforms that are optimal for \sr SemSeg performance. For ObjRec, we find it empirically beneficial to fix geometric transforms to a \texttt{CenterCrop} and use photometric transforms from RandAugment. These choices follow~\citep{hoyer2022mic}. These operations are -- \texttt{AutoContrast}, \texttt{Equalize}, \texttt{Contrast}, \texttt{Brightness}, \texttt{Sharpness}, \texttt{Posterize}, \texttt{Solarize} and \texttt{SolarizeAdd}. We use $(m=30, n=8)$ while sampling operations from this vocabulary for RandAugment. We refer the reader to~\citep{cubuk2020randaugment} for more details.
\label{sec:aug_choices}

\subsection{Training and Implementation Details}
We outline training and implementation details associated with our experiments.
\par \noindent
\textbf{Semantic Segmentation.} For SemSeg, we use GTAV~\citep{sankaranarayanan2018GTA} as the source \syn dataset and Cityscapes~\citep{cordts2016cityscapes} as the target real dataset. We consider two segmentation architectures for our experiments -- DeepLabv2~\citep{chen2017deeplab} (with a ResNet-101~\citep{he2016deep} backbone) and DAFormer~\citep{hoyer2022daformer} (with an MiT-B5~\citep{xie2021segformer} backbone). Both models are initialized from backbones pretrained on ImageNet~\citep{imagenet}. For EntMin~\citep{vu2019advent} (with DAFormer / DeepLabv2), we use SGD as an optimizer with a learning rate of $2.5\times 10^{-4}$ and use $\lambda_{UDA} = 0.001$ as the coefficient of the ``unconstrained'' entropy loss. For HRDA, we use the multi-resolution self-training strategy from~\citep{hoyer2022hrda} -- AdamW~\citep{loshchilov2017decoupled} as the optimizer with a learning rate of $6\times 10^{-5}$ for the encoder and $6\times 10^{-4}$ for the decoder, with a linear learning rate warmup (warmup iterations $1500$; warmup ratio $10^{-6}$), followed by polynomial decay (to an eventual learning rate of $0$). For the teacher-student self-training setup in HRDA, we additionally couple cross-domain mixing with augmentations (DACS~\citep{tranheden2021dacs}) to improve self-training, use the ImageNet feature distance loss~\citep{hoyer2022daformer}, set $\lambda_{UDA} = 1$ and use $\alpha = 0.999$ as the teacher EMA factor. For all SemSeg settings, we use a batch size of $2$ ($2$ source images, $2$ target images) and additionally use rare class sampling (based on source; following~\citep{hoyer2022daformer}) to ensure consistent adaptation improvements across all classes. For all our SemSeg experiments, we feed crops of size $1024\times 1024$ as input to the models, irrespective of the adaptation method. \pst for \augcal is applied to the same crops. We train all segmentation models for $40$k iterations and use the last checkpoint to report results (the \sr adaptation setting does not assume access to labels on real data which prevents selecting ``best-on-target'' checkpoint). For semantic segmentation, following prior work~\citep{wang2022calibrating}, for ECE (and other associated metrics), instead of pooling all pixels of different images into a set, we compute per-image numbers and then average across images. We use the same to compute standard error. We report all metrics (performance, calibration, etc.) as percentages. We use $15$ bins to compute ECE.

\par \noindent
\textbf{Object Recognition.} For ObjRec, we conduct experiments on the VisDA~\citep{peng2017visda} \sr benchmark, using standard source-target splits~\citep{hoyer2022mic} for training. We consider ImageNet~\citep{imagenet} pretrained ResNet-101~\citep{he2016deep} and ViT-B/16~\citep{dosovitskiy2020image} backbones with standard bottleneck (with Linear, BatchNorm and ReLU) and classifier layers. As noted earlier, we use SDAT~\citep{rangwani2022closer} as the adaptation method, which relies on Conditional Domain Adversarial Adaptation (CDAN)~\citep{long2018conditional} and Minimum Class Confusion (MCC)~\citep{jin2020minimum} with SAM~\citep{foret2020sharpness} (smoothness $0.2$) -- all adaptation losses are combined and $\lambda_{UDA}$ is set to $1$. We use SGD as the base optimizer with a learning rate of $2\times 10^{-4}$, with a batch size of $32$ (for both source and target). \pst for \augcal is applied to the raw input \syn images. We train ResNet-101 backbones for $30$ epochs and ViT-B/16 backbones for $15$ epochs and use the last checkpoint to report results (the \sr adaptation setting does not assume access to labels on real data which prevents selecting ``best-on-target'' checkpoint). We report all metrics (performance, calibration, etc.) as percentages. We use $15$ bins to compute ECE. For our key results in the main paper (Tables. 2(b) \& 3(b)), we run experiments across 3 random seeds.

\par \noindent
\textbf{MIC Hyperparameters.} We also use MIC~\citep{hoyer2022mic} with all the discussed \sr adaptation methods since it demonstrably improves performance. For MIC, following prior work~\citep{hoyer2022mic}, we use a masking patch size of $64$, a masking ratio of $0.7$, a loss weight of $1$ and an EMA factor of $0.999$ for the pseudo-label generating teacher.

\par \noindent
\textbf{Compute.} We conduct all object recognition experiments on RTX 6000 GPUs -- every experiment requiring a single GPU. For semantic segmentation, we use one A40 GPU per experiment.

\begin{table}
\setlength{\tabcolsep}{2pt}
\caption{\footnotesize\textbf{Reproducing \sr Adaptation Baselines.} We summarize results from our reproduction of the \sr adaptation baselines (from~\citep{hoyer2022mic}) used in our experiments. We find a slight difference between reported and reproduced adaptation results across models on both (a) GTAV$\to$Cityscapes and (b) VisDA syn$\to$real. $\pm$ indicates standard error.}
\parbox{.5\linewidth}{
\centering
\resizebox{\linewidth}{!}{
\begin{tabular}{lccc}
\toprule
\textbf{Method} & \textbf{Model} & \textbf{Status} & \textbf{mIoU} ($\uparrow$) \\
\midrule
\texttt{1} HRDA + MIC & DeepLabv2 & \citep{hoyer2022mic} &  64.20\\
\texttt{2} HRDA + MIC & DeepLabv2 & Reproduced &  64.05\\
\midrule
\texttt{3} HRDA + MIC & DAFormer & \citep{hoyer2022mic} & 75.90 \\
\texttt{4} HRDA + MIC & DAFormer & Reproduced &  75.56\\
\bottomrule
\end{tabular}}
\caption*{(a) GTAV$\to$Cityscapes.}
}
\hfill
\parbox{.5\linewidth}{
\centering
\resizebox{\linewidth}{!}{
\begin{tabular}{lccc}
\toprule
\textbf{Method} & \textbf{Model} & \textbf{Status} & \textbf{mAcc} ($\uparrow$) \\
\midrule
\texttt{1} SDAT + MIC & ResNet-101 & \citep{hoyer2022mic} &  86.90\\
\texttt{2} SDAT + MIC & ResNet-101 & Reproduced &  81.03\tiny{\std{2.32}}\\
\midrule
\texttt{3} SDAT + MIC & ViT-B & \citep{hoyer2022mic} &  92.80\\
\texttt{4} SDAT + MIC & ViT-B & Reproduced &  92.62\tiny{\std{0.28}}\\
\bottomrule
\end{tabular}}
\caption*{(b) VisDA \sr.}
}
\label{table:reproduced_numbers}
\end{table}

\par \noindent
\textbf{Reproduced Results.} We use open-sourced code for~\citep{hoyer2022mic}\footnote{\href{https://github.com/lhoyer/MIC}{https://github.com/lhoyer/MIC}} and re-run the base adaptation methods for HRDA + MIC and SDAT + MIC at our end to obtain baseline results. In Table.~\ref{table:reproduced_numbers}, we summarize reported and reproduced results for the same methods. 
\label{sec:training_details}

\subsection{Prediction Rejection Ratio (PRR)}
As noted in Sec. 4 of the main paper, in addition to measuring reduced miscalibration, we also assess the extent to which such improvements in calibration are useful and lead to a more reliable model. To measure this, following prior work~\citep{de2023reliability,malinin2019ensemble}, we measure whether confidence scores can reliably guide misclassification detection -- measured via the PRR metric. 

Ideally, in a real-world scenario, given a model, we would like to retrieve all (potentially) samples misclassified by the model based on confidence scores. These samples can then be skipped when the model is used for decision-making (since the model is likely to make incorrect predictions). Since confidence scores, in practice, are imperfect, measuring misclassification detection helps us assess this specific capability of the \sr adapted models. For a model that is less overconfident on incorrect samples (meaning it has reduced miscalibration), this specific ability should naturally be enhanced.

This can be measured using Rejection-Accuracy curves~\citep{de2023reliability,malinin2019ensemble} where we reject samples below a threshold and keep track of accuracy and the fraction of rejected samples. Since such curves are naturally biased towards models that have improved performance, the AUC for such a rejection curve can be normalized by that of an oracle. Additionally, we can subtract a score associated with a random baseline (sorting predictions for filtering in a random order)~\citep{wang2021rethinking}. Finally, we can compare this value (PRR; ranging from -$100$ to $100$; higher is better) for multiple models to assess how reliable underlying confidence scores are.
\label{sec:prr}

\subsection{\augcal Results}
\begin{table}
\setlength{\tabcolsep}{1pt}
\caption{\footnotesize\textbf{\augcal ensures \sr adapted models make accurate, calibrated and reliable predictions.} We find that applying \augcal to multiple \sr adaptation methods across tasks leads to better calibration (ECE, IC-ECE), reduced overconfidence (OC) and improved reliability (PRR) -- all while retaining or improving transfer performance. Highlighted rows are \augcal variants of the base methods. For \augcal, we use \pst as \aug and DCA
as \calib. $\pm$ indicates standard error.}
\vspace{-10pt}
\parbox{.5\linewidth}{
\centering
\resizebox{0.5\columnwidth}{!}{
\begin{tabular}{lccccccc}
\toprule
\multirow{2}{*}{\textbf{Method}} & \multicolumn{1}{c}{\textbf{Perf. ($\uparrow$)}} & & \multicolumn{3}{c}{\textbf{Calibration Error ($\downarrow$)}} & & \multicolumn{1}{c}{\textbf{Reliability ($\uparrow$)}}\\ 
 & mIoU & & ECE & IC-ECE & OC & & PRR \\
\midrule
\multicolumn{1}{l}{\texttt{1} EntMin + MIC} & 47.77 & & 4.93\tiny{\std{0.20}} & 70.09\tiny{\std{0.22}} & 75.80\tiny{\std{0.58}} & & 48.87\tiny{\std{0.45}}\\
\rowcolor{defaultcolor}
\multicolumn{1}{l}{\texttt{2} \quad + \augcal} & \textbf{49.96} & & \textbf{3.21}\tiny{\std{0.16}} & \textbf{67.99}\tiny{\std{0.22}} & \textbf{74.33}\tiny{\std{0.57}} & & \textbf{52.07}\tiny{\std{0.48}}\\
\multicolumn{1}{l}{\texttt{3} HRDA + MIC} & 64.05 & & 3.52\tiny{\std{0.19}} & 78.94\tiny{\std{0.18}} & 86.48\tiny{\std{0.50}} & & 63.20\tiny{\std{0.48}}\\
\rowcolor{defaultcolor}
\multicolumn{1}{l}{\texttt{4} \quad + \augcal}  & 63.95 & & \textbf{2.55}\tiny{\std{0.11}} & \textbf{74.71}\tiny{\std{0.21}} & \textbf{84.13}\tiny{\std{0.52}} & & \textbf{65.37}\tiny{\std{0.50}}\\
\bottomrule
\end{tabular}}
\caption*{(a) GTAV$\to$Cityscapes. (DeepLabv2 R-101).}
}
\hfill
\parbox{.5\linewidth}{
\centering
\resizebox{0.5\columnwidth}{!}{
\begin{tabular}{lccccccc}
\toprule
\multirow{2}{*}{\textbf{Method}} & \multicolumn{1}{c}{\textbf{Perf. ($\uparrow$)}} & & \multicolumn{3}{c}{\textbf{Calibration Error ($\downarrow$)}} & & \multicolumn{1}{c}{\textbf{Reliability ($\uparrow$)}}\\ 
 & mIoU & & ECE & IC-ECE & OC & & PRR \\
\midrule
\multicolumn{1}{l}{\texttt{1} SDAT + MIC} & 81.03\tiny{\std{2.32}} & & 14.32\tiny{\std{1.41}} & 85.06\tiny{\std{1.17}} & 82.77\tiny{\std{0.88}} & & 46.92\tiny{\std{4.21}}\\
\rowcolor{defaultcolor}
\multicolumn{1}{l}{\texttt{2} \quad + \augcal} & \textbf{84.27}\tiny{\std{2.22}} & & \textbf{11.89}\tiny{\std{1.14}} & \textbf{82.98}\tiny{\std{0.87}} & \textbf{81.10}\tiny{\std{0.63}} & & \textbf{53.26}\tiny{\std{3.52}}\\
\bottomrule
\end{tabular}}
\caption*{(b) VisDA \sr. (ResNet-101).}
}
\label{table:aug_cal_key_res_cnn}
\vspace{-20pt}
\end{table}
\begin{table}
\setlength{\tabcolsep}{1pt}
\caption{\footnotesize\textbf{Comparing \augcal with Temperature Scaling for VisDA \sr.} We do a controlled experiment (with an 80-20 split of the VisDA \sr split) to compare \augcal with Temperature Scaling (TS). B = SDAT + MIC (ViT-B). C (clean) and P (\pst augmented) indicate the synthetic ``labeled'' validation splits the temperature was tuned on. For \augcal, we use \pst as \aug and DCA as \calib. $\pm$ indicates standard error.}
\vspace{-10pt}
\parbox{.5\linewidth}{
\centering
\resizebox{0.5\columnwidth}{!}{
\begin{tabular}{lccccccc}
\toprule
\multirow{2}{*}{\textbf{Method}} & \multicolumn{1}{c}{\textbf{Perf. ($\uparrow$)}} & & \multicolumn{3}{c}{\textbf{Calibration Error ($\downarrow$)}} & & \multicolumn{1}{c}{\textbf{Reliability ($\uparrow$)}}\\ 
 & mIoU & & ECE & IC-ECE & OC & & PRR \\
\midrule
\multicolumn{1}{l}{\texttt{1} SDAT + MIC (B)} & 92.24\tiny{\std{0.30}} & & 8.13\tiny{\std{0.36}} & 91.54\tiny{\std{0.05}} & 89.48\tiny{\std{0.43}} & & 63.92\tiny{\std{1.46}}\\
\multicolumn{1}{l}{\texttt{2} B + TS (C)} & 92.24\tiny{\std{0.30}} & & 8.19\tiny{\std{0.26}} & 95.41\tiny{\std{0.08}} & 93.97\tiny{\std{0.20}} & & 66.58\tiny{\std{1.23}}\\
\multicolumn{1}{l}{\texttt{3} B + TS (P)} & 92.24\tiny{\std{0.30}} & & 8.54\tiny{\std{0.40}} & 93.52\tiny{\std{0.36}} & 92.00\tiny{\std{0.42}} & & 64.05\tiny{\std{1.50}}\\
\rowcolor{defaultcolor}
\multicolumn{1}{l}{\texttt{4} B + \augcal} & \textbf{92.68}\tiny{\std{0.17}} & & \textbf{7.24}\tiny{\std{0.11}} & \textbf{90.40}\tiny{\std{0.09}} & \textbf{87.57}\tiny{\std{0.61}} & & \textbf{66.49}\tiny{\std{0.44}}\\
\bottomrule
\end{tabular}}
\caption*{(b) \augcal vs Temperature Scaling.}
}
\hfill
\parbox{.5\linewidth}{
\centering
\resizebox{0.5\columnwidth}{!}{
\begin{tabular}{lccccccc}
\toprule
\multirow{2}{*}{\textbf{Method}} & \multicolumn{1}{c}{\textbf{Perf. ($\uparrow$)}} & & \multicolumn{3}{c}{\textbf{Calibration Error ($\downarrow$)}} & & \multicolumn{1}{c}{\textbf{Reliability ($\uparrow$)}}\\ 
 & mIoU & & ECE & IC-ECE & OC & & PRR \\
\midrule
\multicolumn{1}{l}{\texttt{1} SDAT + MIC (B)} & 92.24\tiny{\std{0.30}} & & 8.13\tiny{\std{0.36}} & 91.54\tiny{\std{0.05}} & 89.48\tiny{\std{0.43}} & & 63.92\tiny{\std{1.46}}\\
\rowcolor{defaultcolor}
\multicolumn{1}{l}{\texttt{2} \quad + \augcal} & \textbf{92.68}\tiny{\std{0.17}} & & \textbf{7.24}\tiny{\std{0.11}} & \textbf{90.40}\tiny{\std{0.09}} & \textbf{87.57}\tiny{\std{0.61}} & & \textbf{66.49}\tiny{\std{0.44}}\\
\multicolumn{1}{l}{\texttt{3} \qquad + TS (C)} & 92.68\tiny{\std{0.17}} & & 8.01\tiny{\std{0.13}} & 94.44\tiny{\std{0.17}} & 92.78\tiny{\std{0.53}} & & 66.10\tiny{\std{0.65}}\\
\multicolumn{1}{l}{\texttt{4} \qquad + TS (P)} & 92.68\tiny{\std{0.17}} & & 8.02\tiny{\std{0.12}} & 94.46\tiny{\std{0.20}} & 92.80\tiny{\std{0.58}} & & 66.05\tiny{\std{0.65}}\\
\bottomrule
\end{tabular}}
\caption*{(b) \augcal + Temperature Scaling.}
}
\label{table:augcal_vs_tscale}
\vspace{-20pt}
\end{table}

\par \noindent
\textbf{$\triangleright$ \augcal results with CNN architectures.} Key results presented in Table. 2 of the main paper use transformer architectures (DAFormer for SemSeg, ViT-B for ObjRec). In Tables.~\ref{table:aug_cal_key_res_cnn} (a) and (b), we verify that \augcal improvements translate to CNN based architectures as well (also discussed in Sec. 5.2 of the paper). We use DeepLabv2 (ResNet-101) for SemSeg and a ResNet-101 based classifier for ObjRec and find that applying \augcal improves or retains performance, reduces miscalibration and improves reliability.

\par \noindent
\textbf{$\triangleright$ Comparing \augcal with Temperature Scaling (TS).} While we focus on ``training-time'' patches to improve \sr calibration, we also conduct an experiment to compare \augcal with ``post-hoc'' temperature scaling (TS) on VisDA.
Specifically, we use 80\% of VisDA \syn images for training models and rest (20\%) for validation and temperature tuning. To ensure a fair comparison, we consider temperature tuning on both "clean" (C) and "\pst augmented" (P) val splits.\footnote{Note that temperature tuning (by definition) only affects calibration and has no impact on performance.} We present these results in Table.~\ref{table:augcal_vs_tscale} (a). We find that irrespective of tuning on C or P, unlike \augcal, TS is ineffective and increases overconfidence and miscalibration. In Table.~\ref{table:augcal_vs_tscale}, we additionally consider temperature scaling on the logits of an \augcal improved \sr model and find that it worsens miscalibration and overconfidence. Note that this is not entirely surprising since TS depends heavily on the ``data-split'' the temperature is tuned on, which in our case is \syn and not \real.

\par \noindent
\textbf{$\triangleright$ Sensitivity to $\lambda_{\textsc{Cal}}$.} For an existing \sr adaptation pipeline, \augcal involves optimizing for an additional calibration loss $\mathcal{L}_{\textsc{Cal}}$ on augmented synthetic images. In Table.~\ref{table:lambda_val_semseg}, we vary the coefficient $\lambda_{\textsc{Cal}}$ (values in the set $\{0.1, 0.5, 1.0, 5.0, 10.0, 20.0, 100.0\}$) for $\mathcal{L}_{\textsc{Cal}}$  and note the effect on adaptation performance, calibration on real data (ECE, IC-ECE) and reliability (PRR). In Table 2 of the main paper, we already note how applying \augcal with $\lambda_{\textsc{Cal}} = 1$ improves over a baseline adaptation method (\md{red} and \ac{blue} rows in Table.~\ref{table:lambda_val_semseg}). We further note that compared to other values, our choice of $\lambda_{\textsc{Cal}} = 1.0$ achieves a balance between adaptation performance, reduced miscalibration and improved reliability. We find that overly high values of $\lambda_{\textsc{Cal}} \geq 5$ can potentially lead to reduced adaptation performance -- $\lambda_{\textsc{Cal}} \geq 5$ significantly raises the scale of $\mathcal{L}_{\textsc{Cal}}$ compared to $\mathcal{L}_{CE}$ and $\mathcal{L}_{UDA}$, which leads to models optimizing for improved calibration at the expense of task performance. Based on our experiments across multiple models, shifts and tasks, we recommend restricting $\lambda_{\textsc{Cal}} < 5$ for \sr adaptation.

\begin{table}
\setlength{\tabcolsep}{2pt}
\caption{\footnotesize\textbf{Sensitivity to $\lambda_{\textsc{Cal}}$ for \augcal on GTAV$\to$Cityscapes.} We vary the value of $\lambda_{\textsc{Cal}}\in\{0.1, 0.5, 1.0, 5.0, 10.0, 20.0, 100.0 \}$ and report the effect on adaptation performance, reduced miscalibration and improved reliability. We find that our choice of \ac{$\lambda_{\textsc{Cal}} = 1$} leads to balanced performance across desired metrics. Rows in \md{red} correspond to the baseline \sr adaptation method without the application of \augcal.}
\parbox{.5\linewidth}{
\centering
\resizebox{\linewidth}{!}{
\begin{tabular}{lccccc}
\toprule
\multirow{2}{*}{\textbf{Value of $\lambda_{\textsc{Cal}}$}} & \multicolumn{1}{c}{\textbf{Perf. ($\uparrow$)}} & \multicolumn{2}{c}{\textbf{Calibration Error ($\downarrow$)}} & \multicolumn{1}{c}{\textbf{Reliability ($\uparrow$)}}\\ 
 & mIoU  & ECE  & IC-ECE  & PRR  \\
\midrule
\rowcolor{red!15}
\texttt{1} No \augcal & 65.71 & 5.34 & 77.73 & 45.93\\
\midrule
\texttt{2} $\lambda_{\textsc{Cal}}=0.1$ & 69.72 & 2.88 & 74.55 & 57.06\\
\texttt{3} $\lambda_{\textsc{Cal}}=0.5$ & 69.27 & 2.71 & 73.54 & 60.32\\
\rowcolor{defaultcolor}
\texttt{4} $\lambda_{\textsc{Cal}}=1.0$ & 70.31 & 3.43 & 72.97 & 62.66\\
\texttt{5} $\lambda_{\textsc{Cal}}=5.0$ & 66.24 & 3.55 & 70.22 & 62.66\\
\texttt{6} $\lambda_{\textsc{Cal}}=10.0$ & 64.27 & 3.28 & 67.63 & 64.55\\
\texttt{7} $\lambda_{\textsc{Cal}}=20.0$ & 55.37 & 4.17 & 65.55 & 63.20\\
\texttt{8} $\lambda_{\textsc{Cal}}=100.0$ & 27.73 & 6.06 & 52.24 & 54.47\\
\bottomrule
\end{tabular}}
\caption*{(a) EntMin + MIC (DAFormer).}
}
\hfill
\parbox{.5\linewidth}{
\centering
\resizebox{\linewidth}{!}{
\begin{tabular}{lccccc}
\toprule
\multirow{2}{*}{\textbf{Value of $\lambda_{\textsc{Cal}}$}} & \multicolumn{1}{c}{\textbf{Perf. ($\uparrow$)}} & \multicolumn{2}{c}{\textbf{Calibration Error ($\downarrow$)}} & \multicolumn{1}{c}{\textbf{Reliability ($\uparrow$)}}\\ 
 & mIoU  & ECE  & IC-ECE  & PRR  \\
\midrule
\rowcolor{red!15}
\texttt{1} No \augcal & 75.56 & 2.86 & 81.92 & 68.91\\
\midrule
\texttt{2} $\lambda_{\textsc{Cal}}=0.1$ & 75.25 & 2.94 & 80.68 & 69.96 \\
\texttt{3} $\lambda_{\textsc{Cal}}=0.5$ & 75.05 & 2.77 & 80.23 & 70.62\\
\rowcolor{defaultcolor}
\texttt{4} $\lambda_{\textsc{Cal}}=1.0$ & 75.90 & 2.45 & 79.09 & 70.35 \\
\texttt{5} $\lambda_{\textsc{Cal}}=5.0$ & 73.80 & 2.25 & 76.61 & 70.54\\
\texttt{6} $\lambda_{\textsc{Cal}}=10.0$ & 71.28 & 2.50 & 75.85 & 69.17 \\
\texttt{7} $\lambda_{\textsc{Cal}}=20.0$ & 62.31 & 2.43 & 73.32 & 68.27 \\
\texttt{8} $\lambda_{\textsc{Cal}}=100.0$ & 31.01 & 7.68 & 73.56 & 56.52\\
\bottomrule
\end{tabular}}
\caption*{(a) HRDA + MIC (DAFormer).}
}
\label{table:lambda_val_semseg}
\end{table}

\label{sec:augcal_res}

\subsection{Qualitative Predictions}
\begin{figure}[t]
  \centering
\includegraphics[width=\textwidth]{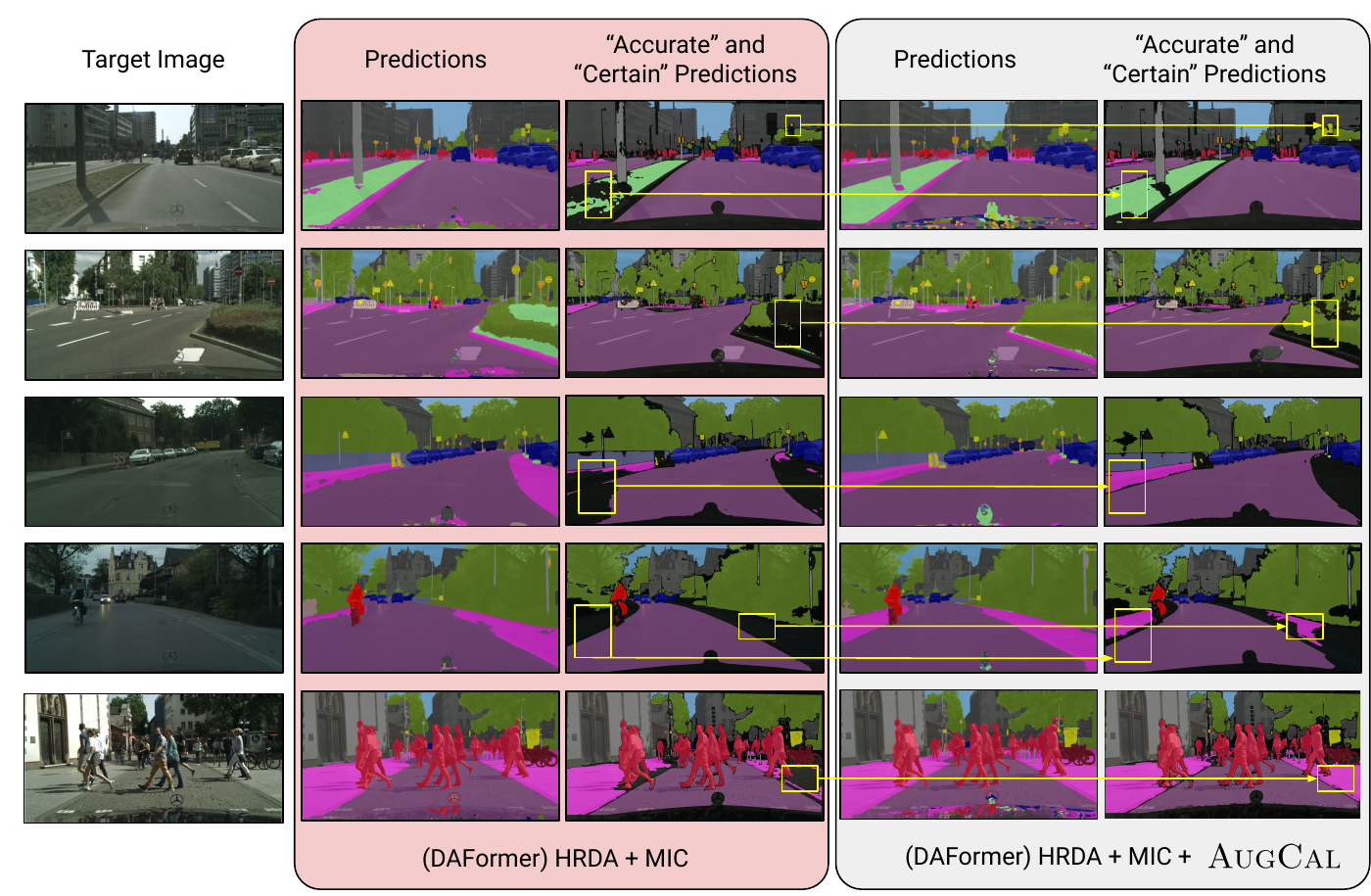}
  \caption{\footnotesize\textbf{\augcal increases the proportion of ``accurate'' and ``certain'' predictions}. For a base DAFormer SemSeg model trained with HRDA + MIC (State-of-the-art) on GTAV$\to$Cityscapes, we show how applying \augcal at training time (right) can improve the proportion of ``accurate'' and ``certain'' (confidence $> 0.95$) predictions over the vanilla adaptation method (left). Regions in black do not satisfy the ``accurate'' and ``certain'' filtering criteria.}
  \label{fig:augcal_hrda_appendix_qual}
\end{figure}

\begin{figure}[t]
  \centering
\includegraphics[width=\textwidth]{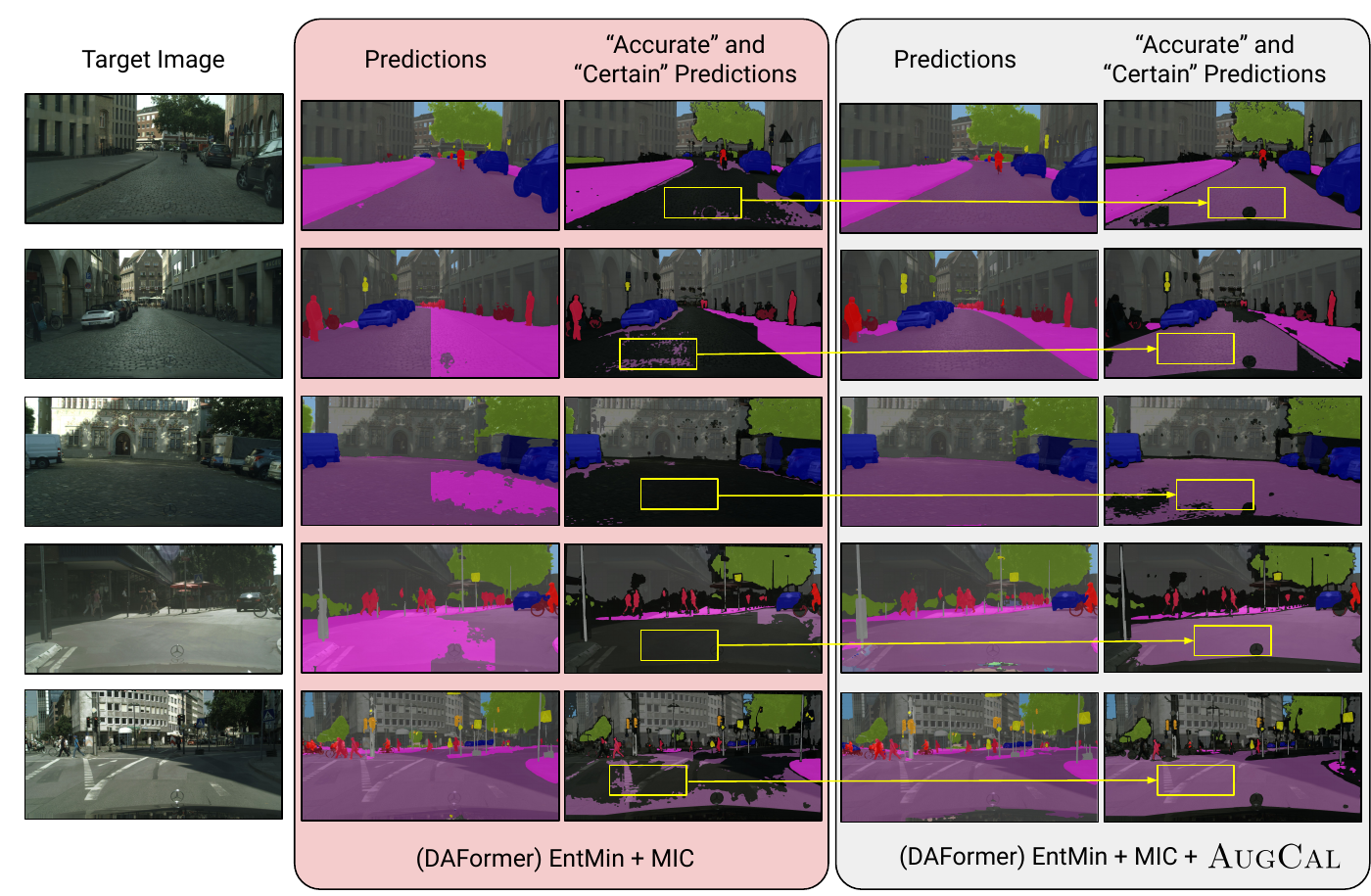}
  \caption{\footnotesize\textbf{\augcal increases the proportion of ``accurate'' and ``certain'' predictions}. For a base DAFormer SemSeg model trained with EntMin + MIC on GTAV$\to$Cityscapes, we show how applying \augcal at training time (right) can improve the proportion of ``accurate'' and ``certain'' (confidence $> 0.95$) predictions over the vanilla adaptation method (left). Regions in black do not satisfy the ``accurate'' and ``certain'' filtering criteria.}
  \label{fig:augcal_emin_appendix_qual}
\end{figure}

In Figures~\ref{fig:augcal_hrda_appendix_qual} and~\ref{fig:augcal_emin_appendix_qual}, we provide more examples to demonstrate how \augcal improves the proportion of ``accurate'' and ``certain'' predictions.
In Fig.~\ref{fig:augcal_hrda_appendix_qual}, where we compare predictions for a (DAFormer) HRDA + MIC model -- w/o \augcal $75.56$ mIoU and w \augcal $75.90$ mIoU.
We find that applying \augcal improves the proportion of ``accurate'' and ``certain'' predictions (confidence $>0.95$) -- see Fig.~\ref{fig:augcal_hrda_appendix_qual}, columns 3 and 5, guided by yellow arrows. For instance, we find that the base model (w/o \augcal) has trouble assigning high-confidence to correct ``sidewalk'' and ``vegetation'' predictions. For EntMin + MIC (w/o \augcal $65.71$ mIoU and w \augcal $70.31$ mIoU) in Fig.~\ref{fig:augcal_emin_appendix_qual}, we find that \augcal is considerably more effective in ensuring high-confidence correct predictions. 
Notably, we find these improvements to be more subtle as the \sr adaptation method itself improves (HRDA + MIC > EntMin + MIC).

\label{sec:qual_pred}

\subsection{\sr Adaptation Methods}
We conduct experiments with three \sr adaptation methods across two tasks. We wanted to assess the compatibility of \augcal with \sr adaptation methods that are competitive and are representative of the broader class of approaches used for \sr transfer. We first cover some background and describe these methods in detail.

In unsupervised domain adaptation (UDA) for \sr settings, we assume access to a labeled (\syn) source dataset $D_S = \{(x^S_i, y^S_i) \}_{i=1}^{|S|}$ and an unlabeled (\real) target dataset $D_T = \{x^T_i\}_{i=1}^{|T|}$. We assume $D_S$ and $D_T$ splits are drawn from source and target distributions $P^S(x,y)$ and $P^T(x,y)$ respectively. At training, we have access to $D = D_S \cup D_T$. We operate in the setting where source and target share the same label space, and discrepancies exist 
only in 
input images. 
The model $\mathcal{M}_{\theta}$ is trained on labeled source images using cross entropy,
\begin{equation}
    \sum_{i=1}^{|S|} \mathcal{L}_{CE} (x_i^S, y_i^S; \theta) = - \sum_{i=1}^{|S|} y_i^S \log p_{\theta}(\hat{y}_i^S | x_i^S) \text{ where }\hat{y}_i^S = \argmax_{y\in\mathcal{Y}} p_{\theta}(y_i^S|x_i^S)
\end{equation}
For object recognition, we represent the labels corresponding to images $x\in\mathbb{R}^{H\times W\times 3}$ as $y\in\mathcal{Y}$ where $\mathcal{Y} = \{1, 2, ..., K\}$. For semantic segmentation, since we make predictions across a vocabulary of classes for every pixel, the corresponding label can be expressed as $y\in\mathcal{Y}^{H\times W}$.

UDA methods additionally optimize for an adaptation objective on labeled source and unlabeled target data ($\mathcal{L}_{UDA}$). The overall learning objective can be expressed as,
\begin{equation}
    \min_{\theta} \underbrace{\sum_{i=1}^{|S|} \mathcal{L}_{CE} (x_i^S, y_i^S; \theta)}_{\text{Source Loss}} + \underbrace{\sum_{i=1}^{|T|} \sum_{j=1}^{|S|} \lambda_{UDA} \mathcal{L}_{UDA} (x_i^T, x_j^S, y_j^S; \theta)}_{\text{Source Target Adaptation Loss}}
    \label{Eqn:sr_adapt}
\end{equation}

Different adaptation methods usually differ in terms of specific instantiations of this objective. As adaptation methods, we use HRDA~\citep{hoyer2022hrda} and EntMin~\citep{vu2019advent} for segmentation and SDAT~\citep{rangwani2022closer} for recognition.

\par \noindent
\textbf{Entropy Minimization (EntMin).} We consider the unconstrained direct entropy minimization approach from~\citep{vu2019advent} as one of the adaptation methods. EntMin builds on top of the assumption the models trained only on source data tend to be under-confident (make high-entropy predictions) on target images. EntMin enforces high prediction certainty on target images by ensuring that the model makes high-confidence predictions on the same. This is realized by minimizing the normalized entropy of target predictions. Specifically, given a target image $x^T$, if the predictive probabilities per-pixel are expressed as $\mathbf{P}^{(h,w,k)}_{x^T}\in[0,1]^{H\times W\times K}$, the adaptation loss on target can be expressed as,
    \begin{equation}
        \mathcal{L}_{UDA} (x^T; \theta) = \frac{1}{HW} \sum_{h,w} \frac{-1}{\log K}\sum_{k=1}^K \mathbf{P}^{(h,w,k)}_{x^T} \log \mathbf{P}^{(h,w,k)}_{x^T} 
    \end{equation}

EntMin can also be viewed as a ``soft-assignment'' version of self-training~\citep{lee2013pseudo,zou2018unsupervised}. We refer the reader to~\citep{vu2019advent} for more details.

\par \noindent
\textbf{Context-Aware High-Resolution Domain-Adaptive Semantic Segmentation (HRDA).} The base adapation method in HRDA~\citep{hoyer2022hrda} is self-training~\citep{zou2018unsupervised,lee2013pseudo}. HRDA additionally introduces components that are beneficial for domain-adaptive semantic segmentation. Specifically, HRDA proposes a multi-resolution framework by relying on (1) a low-resolution \textit{context} crop to learn long-range contextual dependencies and (2) a high-resolution \textit{detail} crop to make detailed and accurate predictions. During adaptation, HRDA fuses predictions from both crops using input dependent attention. For a target image $x^T$, pseudo-labels are obtained from teacher network $\mathcal{M}_{\phi}$ (a moving average of $\mathcal{M}_{\theta}$) as $\mathbf{P}_{x^T}^{(h,w)} = \argmax_k \mathcal{M}_{\phi}(x^T)$. HRDA adapts to target by minimizing cross entropy w.r.t. $\mathbf{P}_{x^T}^{(h,w)}$ as,
\begin{equation}
    \mathcal{L}_{UDA} (x^T; \theta) = -\sum_{h,w} q_{x^T}\mathbf{P}_{x^T}^{(h,w)} \log \mathcal{M}_{\theta}(x^T) \text{ with } q_{x^T} \text{ being a quality estimate for }\mathbf{P}_{x^T}^{(h,w)}
\end{equation}

Instead of ``self''-training on target images, HRDA uses cross-domain mixing (DACS~\citep{tranheden2021dacs}) to obtain pseudo-labels on augmented images. Additionally, components from DAFormer~\citep{hoyer2022daformer} -- rare class sampling and an ImageNet feature distance loss -- are incorporated in the pipeline to facilitate better adaptation. The quality estimate $q_{x^T}$ is computed as the proportion of pixels that have confidence above a specified threshold. This naturally ensures a warmup stage, where a model is first trained only on synthetic images for a few iterations, followed by training on both synthetic and real images. We refer the reader to~\citep{hoyer2022hrda} for more low-level implementation details.

\par \noindent
\textbf{Smooth Domain Adversarial Training (SDAT).} SDAT~\citep{rangwani2022closer} is an adaptation method for object recognition. The underlying adaptation method in SDAT is domain adversarial training (DAT), which involves reducing the discrepancy between source and target image distributions. This is realized by confusing an additional discriminator that is designed to distinguish between source and target samples. While multiple versions of DAT exist, SDAT uses CDAN~\citep{long2018conditional} as the default DAT method. SDAT investigates the loss-landscapes of DAT style methods and notes that smoother loss-landscapes on source data result in improved transfer to target. Consequently, SDAT proposes optimizing for smoother loss landscapes on labeled source data by modifying the supervised $\mathcal{L}_{CE}$ loss as,
\begin{equation}
    \sum_{i=1}^{|S|} \mathcal{L}_{CE} (x_i^S, y_i^S; \theta) =  \sum_{i=1}^{|S|} y_i^S \max_{||\epsilon|| \leq \rho} \log p_{\theta + \epsilon}(\hat{y}_i^S | x_i^S)
\end{equation}

Perturbing the weights of the network $\theta$ by $\epsilon$ in some neighborhood $\rho$ ensures lower loss values in that neighborhood, thereby encouraging a smoother loss landscape. In practice, this is realized by using sharpness aware minimization (SAM)~\citep{foret2020sharpness} to update model parameters. We refer the reader to~\citep{rangwani2022closer,long2018conditional} for more details on the UDA losses used.

\par \noindent
\textbf{Masking Image Consistency (MIC).} MIC~\citep{hoyer2022mic} is a general technique applicable to any existing adaptation method to improve context utilization while making predictions. MIC involves a teacher-student self-training setup where pseudo-labels are obtained from a weakly augmented target sample (via the teacher). Student predictions for a masked image are forced to match the obtained pseudo-labels. MIC relies on the recent success of masking as an auxiliary task~\citep{he2022masked} but instead of reconstruction, MIC sets up a prediction consistency task. This simple addition to an existing adaptation pipeline leads to considerable improvements across multiple tasks and shifts. A ``masked'' image for MIC is obtained by dividing the original image into patches and randomly masking a subset of the same. Similar to HRDA~\citep{hoyer2022hrda}, the MIC (consistency) loss is multiplied by a quality estimate of the pseudo-labels. We refer the reader to~\citep{hoyer2022mic} for more details on MIC. 
\label{sec:sr_adapt}

\subsection{Assets and Licences}
The assets used in this work can be grouped into three categories -- Datasets, Code Repositories and Dependencies. We discuss source and licences for each of these below.

\par \noindent
\textbf{Datasets.} For semantic segmentation, we use the GTAV~\citep{richter2016playing} and Cityscapes~\citep{cordts2016cityscapes} datasets.
Code used to extract densely annotated images from the GTAV game is distributed under the MIT license.\footnote{\href{https://bitbucket.org/visinf/projects-2016-playing-for-data/src/master/}{https://bitbucket.org/visinf/projects-2016-playing-for-data/src/master/}} The Cityscapes' license agreement dictates that the dataset is made freely available to academic and non-academic entities for non-commercial purposes such as academic research, teaching, scientific publications, or personal experimentation and that permission to use the data is granted under certain conditions.\footnote{\href{https://www.cityscapes-dataset.com/license/}{https://www.cityscapes-dataset.com/license/}} For object recognition, we use the VisDA syn$\to$real~\citep{peng2017visda} benchmark. The VisDA-C development kit on github does not have a license associated with it, but it does include a Terms of Use, which primarily states that the dataset must be used for non-commercial and educational purposes only.\footnote{\href{https://github.com/VisionLearningGroup/taskcv-2017-public/tree/master/classification}{https://github.com/VisionLearningGroup/taskcv-2017-public/tree/master/classification}}

\par \noindent
\textbf{Code Repositories.} For our experiments, apart from code that we wrote ourselves, we build on top of the open-sourced codebase for MIC~\citep{hoyer2022mic}\footnote{\href{https://github.com/lhoyer/MIC}{https://github.com/lhoyer/MIC}} repository.
MIC is distributed under the MIT License. 

\par \noindent
\textbf{Dependencies.} We use Pytorch~\citep{NEURIPS2019_9015} as the deep-learning framework for all our experiments. Pytorch, released by Facebook, is distributed under a Facebook-specific license.\footnote{\href{https://github.com/pytorch/pytorch/blob/master/LICENSE}{https://github.com/pytorch/pytorch/blob/master/LICENSE}}

\label{sec:assets}

\end{document}